\documentclass[11pt]{article}

\usepackage{fullpage}
\usepackage{graphicx}
\usepackage{amsfonts}
\usepackage{booktabs}
\usepackage{float}
\usepackage{amsmath, amssymb, latexsym, amscd, amsthm, epsfig}
\usepackage{multirow}
\usepackage{array}
\usepackage{tabu}
\usepackage{verbatim}
\usepackage[table]{xcolor}
\usepackage{adjustbox}
\usepackage{subcaption}
\usepackage{epstopdf}
\usepackage{makeidx}
\usepackage{tikz}
 \usepackage{hyperref}
\usepackage{listings}
\usepackage[ruled]{algorithm2e}
\usepackage{setspace}
\usepackage{rotating}
\usepackage{bbm}
\usepackage[T1]{fontenc}
\usepackage[utf8]{inputenc}
\usepackage{authblk}

\def\CpR{\textsuperscript{\sffamily\textregistered} }
\def\tt#1{\texttt{#1}}

\def\RefFig#1{Figure~\ref{#1}}
\def\RefTab#1{Table~\ref{#1}}

\def\RefSec#1{Section~\ref{#1}}

\def\RefList#1{Listing~\ref{#1}}

\def\SMLtb{Statistic and Machine Learning Toolbox\textsuperscript{\tiny \texttrademark} }
\def\NNtb{Neural Network Toolbox\textsuperscript{\tiny \texttrademark} }
\def\PCtb{Parallel Computing Toolbox\textsuperscript{\tiny \texttrademark} }
\def\Cpr{\textsuperscript{\tiny \textregistered} }
\def\MTL{{\scshape Matlab}}

\lstdefinestyle{ourMatlab}{ %
  backgroundcolor=\color{white},   
  basicstyle=\footnotesize,        
  breakatwhitespace=false,         
  breaklines=true,                 
  captionpos=b,                    
  commentstyle=\color{gray},    
  deletekeywords={...},            
  escapeinside={\%*}{*)},          
  extendedchars=true,              
  frame=single,	                   
  keepspaces=true,                 
  keywordstyle=\color{blue},       
  language=Octave,                 
  otherkeywords={*,...},           
  numbers=left,                    
  numbersep=5pt,                   
  numberstyle=\tiny\color{gray}, 
  rulecolor=\color{black},         
  showspaces=false,                
  showstringspaces=false,          
  showtabs=false,                  
  stepnumber=2,                    
  stringstyle=\color{black},     
  tabsize=2,	                   
  title=\lstname                   
}

\lstdefinestyle{ourLua}{ %
  backgroundcolor=\color{white},   
  basicstyle=\footnotesize,        
  breakatwhitespace=false,         
  breaklines=true,                 
  captionpos=b,                    
  commentstyle=\color{gray},    
  deletekeywords={...},            
  escapeinside={\%*}{*)},          
  extendedchars=true,              
  frame=single,	                   
  keepspaces=true,                 
  keywordstyle=\color{blue},       
  language={[5.0]Lua},                 
  otherkeywords={*,...},           
  numbers=left,                    
  numbersep=5pt,                   
  numberstyle=\tiny\color{gray}, 
  rulecolor=\color{black},         
  showspaces=false,                
  showstringspaces=false,          
  showtabs=false,                  
  stepnumber=2,                    
  stringstyle=\color{black},     
  tabsize=2,	                   
  title=\lstname                   
}

\lstdefinestyle{ourPython}{ %
  backgroundcolor=\color{white},   
  basicstyle=\footnotesize,        
  breakatwhitespace=false,         
  breaklines=false,                 
  captionpos=b,                    
  commentstyle=\color{gray},    
  deletekeywords={...},            
  escapeinside={\%*}{*)},          
  extendedchars=true,              
  frame=single,	                   
  keepspaces=true,                 
  keywordstyle=\color{blue},       
  language=python,                 
  otherkeywords={*,...},           
  numbers=left,                    
  numbersep=5pt,                   
  numberstyle=\tiny\color{gray}, 
  rulecolor=\color{black},         
  showspaces=false,                
  showstringspaces=false,          
  showtabs=false,                  
  stepnumber=2,                    
  stringstyle=\color{black},     
  tabsize=2,	                   
  title=\lstname                   
}

\def\mcode{\lstset{style=ourMatlab}}
\def\lcode{\lstset{style=ourLua}}
\def\pcode{\lstset{style=ourPython}}

\newcommand*{\affaddr}[1]{#1} 
\newcommand*{\affmark}[1][*]{\textsuperscript{#1}}
\newcommand*{\email}[1]{\texttt{#1}}

\title{{\Huge {\bfseries  Neural Networks for Beginners}} \\ {\Large {\bfseries A fast implementation in \MTL, Torch, TensorFlow}}\vspace{0.5cm}}
\author{F. Giannini\affmark[1], V. Laveglia\affmark[1,2], A. Rossi\affmark[1,3]\thanks{Corresponding Author}, D. Zanca\affmark[1,2], A. Zugarini\affmark[1]\vspace{0.3cm}\\
\affaddr{\affmark[1]DIISM, University of Siena, Siena, Italy}\\
\affaddr{\affmark[2]DINFO, University of Florence, Florence, Italy}\\
\affaddr{\affmark[3]Fondazione Bruno Kessler, Trento, Italy}\\
\vspace{0.5cm}
\email{rossi111@unisi.it}\\
\email{\{giannini7, andrea.zugarini\}@student.unisi.it}\\
\email{\{vincenzo.laveglia, dario.zanca\}@unifi.it}
}

\begin{document}

\maketitle
\vfill 

\section*{What is this report about?} 

This report provides an introduction to some Machine Learning tools within the most common development environments. It mainly focuses on practical problems, skipping any theoretical introduction. It is oriented to both students trying to approach Machine Learning and experts looking for new frameworks.

The dissertation is about Artificial Neural Networks (ANNs~\cite{bishop:2006:PRML,rumelhart1988learning}), since currently is the most trend topic, achieving state of the art performance in many Artificial Intelligence tasks. After a first individual introduction to each framework, the setting up of general practical problems is carried out simultaneously, in order to make the comparison easier.

Since the treated argument is widely studied and in continuos and fast growing, we pair this document with an on-line documentation available at the Lab GitHub repository~\cite{gitlab} which is more dynamic and we hope to be kept updated and possibly enlarged.

\newpage

\tableofcontents

\newpage

\newpage

\section{\MTL: a unified friendly environment}

\subsection{Introduction}

\MTL\CpR \cite{MATLAB:2015} is a very powerful instrument allowing an easy and fast handling of almost every kind of numerical operation, algorithm, programming and testing. The intuitive and friendly interactive interface makes it easy to manipulate, visualize and analyze data. The software provides a lot of mathematical built-in functions for every kind of task and an extensive and easily accessible documentation. It is mainly designed to handle matrices and, hence, almost all the functions and operations are vectorized, i.e. they can manage scalars, as well as vectors, matrices and (often) tensors. For these reasons, it is more efficient to avoid loops cycles (when possible) and to set up operations exploiting matrices multiplication. 

In this document we just show some simple Machine Learning related instruments in order to start playing with ANNs. We assume a basic-level knowledge and address to official documentation for further informations. For instance, you can find informations on how to obtain the software from the official web site\footnote{\url{https://ch.mathworks.com/products/matlab.html?s_tid=hp_products_matlab}}. Indeed, the license is not for free and even if most universities provide a classroom license for students use, maybe could not be possible to access to all the current packages. In particular the \SMLtb and the \NNtb provide a lot of built-in functions and models to implement different ANNs architectures suitable to face every kind of task. The access to both the tools is fundamental in the prosecution, even if we refer to some simple independent examples. The most easy to-go is the \texttt{nnstart} function, which activates a simple GUI guiding the user trough the definition of a simple 2-layer architecture. It allows either to load available data samples or to work with customize data (i.e. two matrices of input data and correspondent target), train the network and analyze the results (Error trend, Confusion Matrix, ROC, etc.). However, more functions are available for specific tasks. For instance, the function \tt{patternnet} is specifically designed for pattern recognition problems, 
\tt{newfit} is suitable for regression, whereas \tt{feedforwardnet} is the most flexible one and allows to build very customized and complicated networks. All the versions are implemented in a similar way and the main options and methods apply to everyone. In the next section we show how to manage customizable architectures starting to face very basic problems. Detailed informations can be find in a dedicated section of the official site\footnote{\url{http://ch.mathworks.com/help/nnet/getting-started-with-neural-network-toolbox.html}}.

\subsubsection*{CUDA\Cpr computing}

GPU computing in \MTL  requires the \PCtb and the CUDA\Cpr installation on the machine. Detailed informations on how to use, check and set GPUs devices can be found in GPU computing official web page\footnote{\url{https://ch.mathworks.com/help/nnet/ug/neural-networks-with-parallel-and-gpu-computing.html}}, where issues on Distributed Computing CPUs/GPUs are introduced too. However, basic operations with graphical cards should in general be quite simple. Data can be moved to the GPU hardware by the function \tt{gpuArray}, then back to the CPU by the function \tt{gather}. When dealing with ANNs, a dedicated function \tt{nndata2gpu} is provided, organizing tensors (representing a dataset) in a efficient configuration on the GPU, in order to speed up the computation. An alternative way is to carry out just the training process in the GPU by the correspondent option of the function \tt{train} (which will be describe in details later). This can be done directly by passing additional arguments, in the \emph{Name,Values} pair notation, the option \tt{'useGPU'} and the value \tt{'yes'}:

\mcode
\begin{lstlisting}
   nn = train(nn, ... , 'useGPU','yes')
\end{lstlisting}

\subsection{Setting up the XOR experiment} 

The XOR is a well-known classification problem, very simple and effective in order to understand the basic properties of many Machine Learning algorithms. Even if writing down an efficient and flexible architecture requires some language expertise, a very elementary implementation can be found in the \MTL section of the GitHub repository\footnote{Available at \url{https://github.com/AILabUSiena/NeuralNetworksForBeginners/tree/master/matlab/2layer}.} of this document. It is not suitable to face real tasks, since no customizations (except for the number of hidden units) are allowed, but can be useful just to give some general tips to design a personal module. The code we present is basic and can be easily improved, but we try to keep it simple just to understand fundamental steps. As we stressed above, we avoid loops exploiting the \MTL efficiency with matrix operations, both in forward and backward steps. This is a key point and it can substantially affects the running time for large data.
\\

\subsubsection*{Initialization}
Here below, we will see how to define and train more efficient architectures exploiting some built-in functions from the \NNtb. Since we face the XOR classification problem, we sort out our experiments by using the function \tt{patternnet}. To start, we have to declare an object of kind \tt{network} by the selected function, which contains variables and methods to carry out the optimization process. The function expects two optional arguments, representing the number of hidden units (and then of the hidden layers) and the back-propagation algorithm to be exploited during the training phase. The number of hidden units has to be provided as a single integer number, expressing the size of the hidden layer, or as an integer row vector, whose elements indicate the size of the correspondent hidden layers. The command:

\begin{lstlisting}
   nn = patternnet(3)
\end{lstlisting}

\noindent creates on object named \tt{nn} of kind \tt{network}, representing a 2-layer ANN with 3 units in the single hidden layer. The object has several options, which can be reached by the dot notation $object.property$ or explore by clicking on the interactively visualization of the object in the \MTL Command Window, which allows to see all the available options for each property too. The second optional parameter selects the training algorithm by a string saved in the \tt{trainFcn} property, which in the default case takes the value '$trainscg$' (Scaled Conjugate Gradient Descent methods). The network object is still not fully defined, since some variables will be adapted to fit the data dimension at the calling of the function \tt{train}. However, the function \tt{configure}, taking as input the object and the data of the problem to be faced, allows to complete the network and set up the options before the optimization starts.

\subsubsection*{Dataset}

Data for ANNs training (as well as for others available Machine Learning methods) must be provided in matrix form, storing each sample column-wise. For example data to define the XOR problem can be simply defined via an input matrix $X$ and a target matrix $Y$ as:

\begin{lstlisting}
   X = [0, 0, 1, 1; 0, 1, 0, 1]
   Y = [0, 1, 1, 0]
\end{lstlisting}

\noindent \MTL expects targets to be provided in $0/1$ form (other values will be rounded). For 2-class problem targets can be provided as a row vector of the same length of the number of samples. For multi-class problem (and as an alternative for 2-class problem too) targets can be provided in the \emph{one-hot encoding} form, i.e. as a matrix with as many columns as the number of samples, each one composed by all $0$ with only a $1$ in the position indicating the class.

\subsubsection*{Configuration}

Once we have defined data, the network can be fully defined and designed by the command:

\begin{lstlisting}
   nn = configure(nn,X,Y)
\end{lstlisting}

\noindent For each layer, an object of kind \tt{nnetLayer} is created and stored in a \emph{cell} array under the field \tt{layers} of the network object. The number of connections (the weights of the network) for each units corresponds to the layer input dimension. The options of each layer can be reached by the dot notation $object.$~\tt{layer\{}$numberOfLayer$\tt{\}}.$property$. The field \tt{initFcn} contains the weights initialization methods. The activation function is stored in the \tt{transferFcn} property. In the hidden layers the default values is the '$tansig$' (Hyperbolic Tangent Sigmoid), whereas the output layers has the '$logsig$' (Logistic Sigmoid) or the '$softmax$' for 1-dimensional and multi-dimensional target respectively. The '$crossentropy$' penalty function is set by default in the field \tt{performFcn}. At this point, the global architecture of the network can be visualized by the command:
\begin{lstlisting}
   view(nn)
\end{lstlisting}

\subsubsection*{Training}

\noindent The function \tt{train} itself makes available many options (as for instance \tt{useParallel} and \tt{useGPU} for heavy computations) directly accessible from its interactive help window. However, it can take as input just the network object, the input and the target matrices. The optimization starts by dividing data in Training, Validation and Test sets. The splitting ratio can be changed by the options \tt{divideParam}. In the default setting, data are randomly divided, but if you want for example to decide which data are used for test, you can change the way the data are distributed by the option \tt{divideFcn}\footnote{Click on \tt{divideFcn} property from the \MTL Command Window visualization of your object to see the available methods.}. In this case, because of the small size of the dataset, we drop validation and test by setting:
 
\begin{lstlisting}
   nn.divideFcn = ''
\end{lstlisting}

\noindent In the following code, we set the training function to the classic gradient descent method '$traingd$', we deactivate the training interactive GUI by \tt{nn.trainParam.showWindow} (boolean) and activate the printing of the training state in the Command Window by \tt{nn.trainParam.showCommandLine} (boolean). Also the \emph{learning rate} is part of the \tt{trainParam} options under the fields \tt{lr}.

\begin{lstlisting}
  nn.trainFcn = 'traingd'
  nn.trainParam.showWindow = 0
  nn.trainParam.showCommandLine = 1
  nn.trainParam.lr = 0.01
\end{lstlisting}

\noindent Training starts by the calling:

\begin{lstlisting}
  [nn, tr] = train(nn, X, Y)
\end{lstlisting}

\noindent this generates a printing, ending in this case with:

\vspace{0.3cm}
\includegraphics{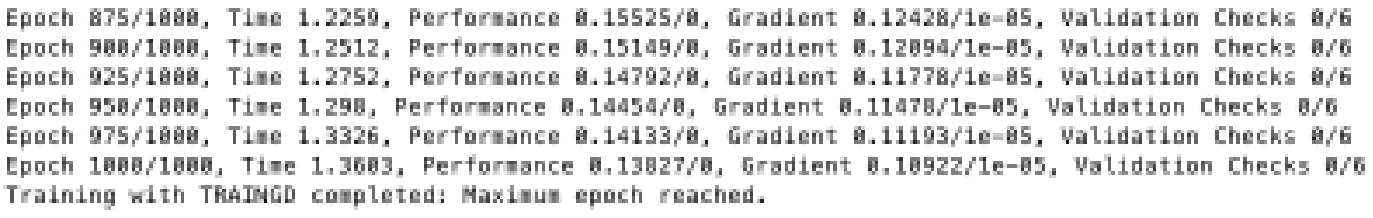}

\noindent This indicates that the training stops after the max number of epoch is reached (which can be set by options $object.$~\tt{trainParam.epochs}). Each column shows the state of one of the stopping criterions used, which we will analyze in details in the next section. The output variable $tr$ stores the training options. The fields \tt{perf}, \tt{vperf} and \tt{tperf} contain the performance of the network evaluated at each epoch on the Training, Validation and Test sets respectively (the last two are \tt{NaN} in this case), which can be used for example to plot performances. If we pass data organized in a single matrix, the function will exploit the full batch learning method accumulating gradients overall the training set. To set a mini-batch mode, data have to be manually split in sub-matrix with the same number of column and organized in a cell array. However, let us consider for a moment a general data set composed by $N$ samples in the features space $\mathbb{R}^D$ with a target of dimension $C$, so that $X \in \mathbb{R}^{D\times N}$ and $Y \in \mathbb{R}^{C\times N}$. All the mini-batches have to be of the same size $b$, so that it is in general convenient to choose the batch size to be a factor of $N$. In this case, we can generate data for the training function organizing the input and target in the correspondent \tt{cell-array} by: 

\begin{lstlisting}
N = size(X,2); % number of samples
n_batch = N/batchsize; % number of batches

input{n_batch} = []; % input cell-array initialization
target{n_batch} = []; % target cell-array initialization

p = randperm(N); % generating a random permutated index for data shuffling
X = X(:,p); % samples permutation
Y = Y(:,p); % target permutaion

for i=1:n_batch
  input{i} = X(:,(1:batchsize)+(i-1)*batchsize); 
  target{i} = Y(:,(1:batchsize)+(i-1)*batchsize);
end
\end{lstlisting}
However, in order to perform a pure \emph{Stochastic Gradient Descent} optimization, in which the ANNs parameters are updated for each sample, the training function \tt{'trains'} have to be employed skipping to split data previously. A remark has to be done since this particular function does not support the GPU computing.

The network (trained or not) can be easily evaluated on data by passing the input data as argument to a function named as the network object. Performance of the prediction with respect to the targets can be evaluated by the function \tt{perform} according to the correspondent loss function option $object.$~\tt{performFcn}:
\begin{lstlisting}
  f = nn(X)
  perform(nn, Y, f)
\end{lstlisting}

\subsection{Stopping criterions and Regularization }\label{regmatlab}

Early stopping is a well known procedure in Machine Learning to avoid overfitting and improve generalization. Routines from \NNtb use different kind of stopping criterions regulated by the network object options in the field \tt{trainParam}. Arbitrarily methods are based on the number of epochs (\tt{epochs}) and the training time (\tt{time}, default $Inf$). A criterion based on the training set check the loss  ($object.$~\tt{trainParam.goal}, default = 0) or the parameters gradients ($object.$~\tt{trainParam.min\_grad}, default = $10^{-6}$) to reach a minimum threshold. A general \emph{early stopping} method is implemented by checking the error on the validation set and interrupting training when validation error does not improve for a number of consecutive epochs given by \tt{max\_fail} (default = 6).

Further regularization methods can be configured by the property \tt{performParam} of the network object. The field \tt{regularization} contains the weight (real number in $[0,1]$) balancing the contribution of a term trying to minimizing the norm of the network weights versus the satisfaction of the penalty function. However, the network is designed to mainly rely on the validation checks, indeed regularization applies only to few kind of penalties and the default weight is 0.

\subsection{Drawing separation surfaces} 

\begin{figure}[H]
 \hspace{-0.6cm}
 \begin{subfigure}{\textwidth}
 	\begin{subfigure}{.45\textwidth}
   		\includegraphics[scale=0.45]{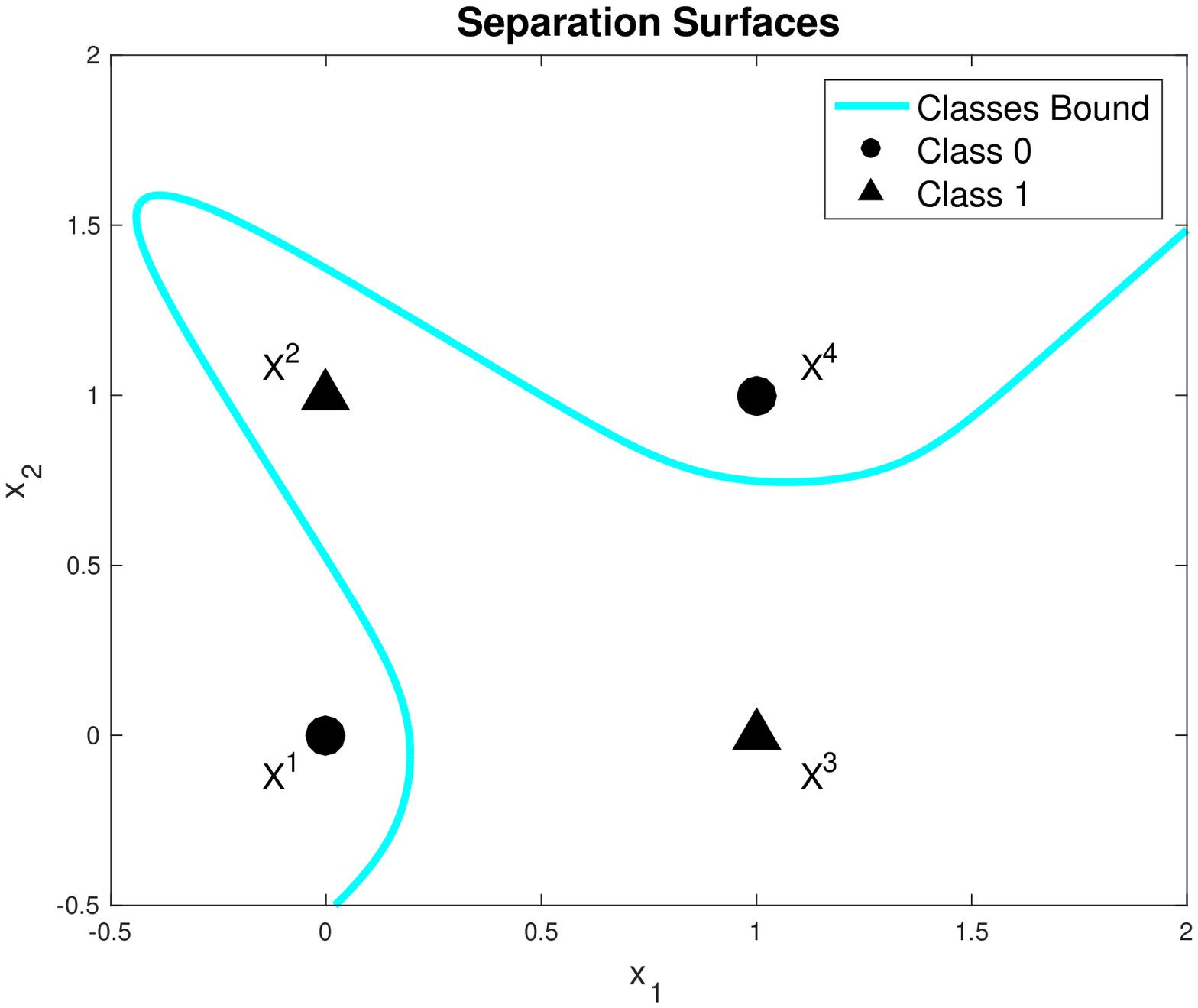}
  		\caption{2-layers, 3 Hidden Units}\label{fh3}
	\end{subfigure}
 \hspace{.05\textwidth}
	\begin{subfigure}{.45\textwidth}
	\vspace{0.3cm}
 		\includegraphics[scale=0.45]{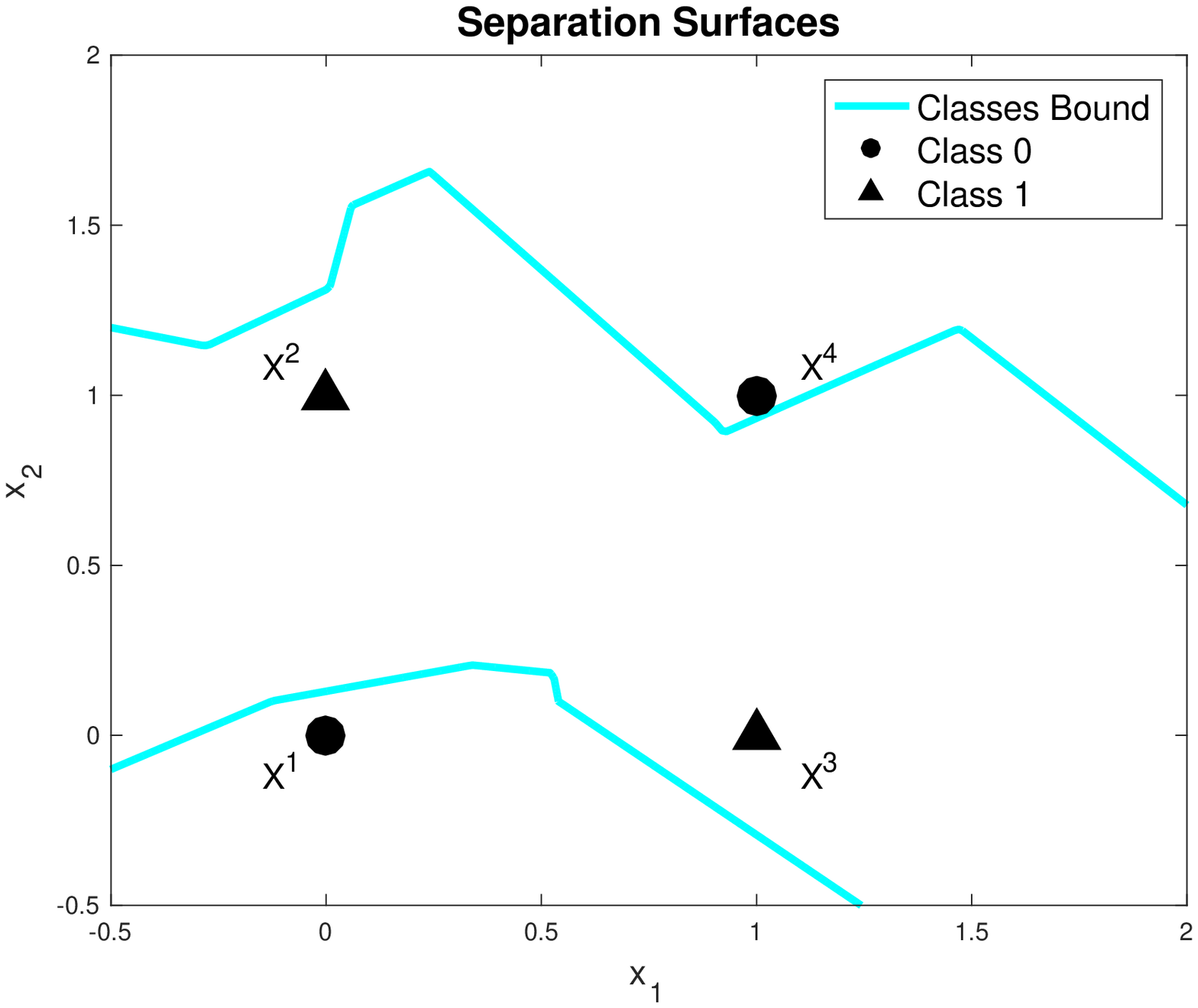}
 		 \caption{4-layers, 5, 3 and 2 Hidden Units, ReLU for all the activations}\label{fh10}
 	\end{subfigure}
\end{subfigure}
\caption{Separation surfaces on the XOR classification task.}\label{sepsurf}
 \end{figure}

When dealing with low dimensional data (as in the XOR case), can be useful to visualize the prediction of the network directly in the input space. For this kind of task, \MTL makes available a lot of built-in functions with many options for interactive data visualization. In this section, we will show the main functions useful to realize customized separation surfaces learned by an ANN with respect to some specific experiments. We briefly add some comments for each instruction, referring to the suite help for specific knowledge of each single function. The network predictions will be evaluated on a grid of the input space, generated by the \MTL function \tt{meshgrid}, since the main functions used for the plotting (\tt{contour} or, if you want a color surface \tt{pcolor}) require as input three matrices of the same dimensions expressing, in each correspondent element, the coordinates of a 3-D point (which in our case will be first input dimension, second input dimension and prediction). Once we trained the network described until now, the boundary for the 2-classes separation showed in \RefFig{fh3} is generated by the code in \RefList{matxorsepsurf}, whereas in \RefFig{fh10} we report the same evaluation after the training of a 4-layers network using 5, 3 and 2 units in the first, second and third hidden layers respectively, each one using the \emph{ReLU} as activation (\tt{'poslin'} in \MTL). This new network can be defined by:

\begin{lstlisting}[caption=Drawing separation surfaces, label=matxorsepsurf]
n = patternnet([5,3,2]); % 3+output layers network initialization
nn = configure(n,X,Y); % network configuration
nn.trainFcn = 'traingd'; % setting optimization function
nn.divideParam.trainRatio = 1; % setting data splitting ratios (illustrative)
nn.divideParam.valRatio = 0;
nn.divideParam.testRatio = 0;
nn.trainParam.showCommandLine = 1;
nn.trainParam.lr = 0.01;
nn.layers{1}.transferFcn = 'poslin'; % setting the activation layer-wise
nn.layers{2}.transferFcn = 'poslin';
nn.layers{3}.transferFcn = 'poslin';
\end{lstlisting}

\begin{lstlisting}
% % % %  Plotting Separation Surface

% generating input space grid
[xp1,xp2] = meshgrid(-0.5:.01:2,-0.5:.01:2); 
% network evaluation on the gridding (reshaped to fit network input dimension)
f = nn([xp1(:)';xp2(:)']);
% reshaping prediction in correspondent matrix form
f = reshape(f,size(xp1,1),[]);

% drawing separation surfaces
contour(xp1,xp2,f,[.5,.5],'LineWidth',3,'Color','c'); 
hold on;

% drawing data points 
scatter(X(1,[1,4]),X(2,[1,4]),200,'o','filled','MarkerEdgeColor','k',...
'MarkerFaceColor','k','LineWidth',2);
scatter(X(1,[2,3]),X(2,[2,3]),200,'^','filled','MarkerEdgeColor','k',...
'MarkerFaceColor','k','LineWidth',2);

axis([-0.5,2,-0.5,2]); % setting axis bounds

% labeling data points
c = {'X^1','X^2','X^3','X^4'}; % labels
dx = [-.15, -.15, .1, .1]; % labels horizontal translation wrt points
dy = [-.1, .1, -.1, .1]; % labels vertical translation wrt points
text(X(1,:)+dx, X(2,:)+dy, c, 'FontSize',14); % showing labels as text

% plot labels
xlabel('x_1','FontSize',14)
ylabel('x_2','FontSize',14)
title('Separation Surfaces','FontSize',16);

h = legend({'Classes Bound','Class 0','Class 1'},'Location','NorthEast');

set(h,'FontSize',14);
\end{lstlisting}


\newpage

\section{Torch and Lua environment} \label{secTorch}
\lcode

\subsection{Introduction}
Torch7 is an easy to use and efficient scientific computing framework, essentially oriented to Machine Learning algorithms. The package is written in C which guarantees an high efficiency. However, a completely interaction is possible (and usually convenient) by the \emph{LuaJIT} interface, which provides a fast and intuitively scripting language. Moreover, it contains all the libraries necessary for the integration with the CUDA\Cpr environment for GPU computing. At the moment of writing it is one of the most used tool for prototyping ANNs of any kind of topology.
Indeed, there are many packages, constantly updated and improved by a large community, allowing to develop almost any kind of architectures in a very simple way.

Informations about the installation can be found at the \emph{getting started} section of the official site\footnote{\url{http://torch.ch/docs/getting-started.html}}. The procedure is straightforward for \emph{UNIX} based operative systems, whereas is not officially supported for Windows, even if an alternative way is provided\footnote{\url{https://github.com/torch/torch7/wiki/Windows}}. If CUDA\Cpr is already installed, also the packages \tt{cutorch} and \tt{cunn} will be added automatically, containing all the necessary utilities to deal with Nvidia GPUs.


\subsection{Getting started} 

\subsubsection{Lua}
Lua, in torch7, acts as an interface for C/CUDA routines. A programmer, in most of the cases, will not have to worry about C functions. Therefore, we explain here only how the Lua language works, because is the only one necessary to deal with Torch. It is a scripting language with a syntax similar to Python and semantic close to Javascript. A variable is considered as \emph{global} by default. The local declaring, which is usually recommended, require the explicit declaration by placing the keyword \tt{local} before the name fo the variable. Lua has been chosen over other scripting languages, such as Python, because is the fastest one, a crucial feature when dealing with large data and complex programs, as common in Machine Learning.

There are seven native types in lua: \tt{nil, boolean, number, string, userdata, function } and \tt{table}, even if most of the Lua power is related to the last one.
A \tt{table} behaves either as an hash map (general case) or as an array (which have the 1-based indexing as in \MTL and Python). The table will be considered as an array when contains only numerical keys, starting from the value $1$. Any other complex structure such as classes, are built from it (formally defined as a \emph{Metatable}).
%
%
%

A detailed documentation on Lua can be find at the official webpage\footnote{Lua $5.1$ reference manual is available here: \url{https://www.lua.org/manual/5.1/}}, however, an essential and fast introduction can be found at \url{http://tylerneylon.com/a/learn-lua/}.

\subsubsection{Torch enviroment}\label{torchintro}
Torch extends the capabilities of the Lua \tt{table} implementing the \tt{Tensor} class. Many Matlab-like functions\footnote{\url{http://atamahjoubfar.github.io/Torch_for_Matlab_users.pdf}} are provided in order to initialize and manipulate tensors in a concise fashion. Most commons are reported in \RefList{torchtens}. 

\begin{lstlisting}[caption= Example of torch tensor basic usages,label=torchtens]
local t1 = torch.Tensor() -- no dimension tensor constructor
local t2 = torch.Tensor(4,3) -- 4x3 empty tensor
local t3 = torch.eye(3,5) -- 3x5 1-diagonal matrix
t2:fill(1) -- fill the matrix with the value 1
t1 = torch.mm(t2,t3) -- assign to t1 the result of matrix multiplication between t2 and t3 
t1[1][2] = 5 -- assign 5 to the element in first row and second column 
\end{lstlisting} 

All the provided packages are developed following a strong modularization, which is a crucial feature to keep the code stable and dynamic. Each one provides several already built-in functionalities, and all of them can be easily imported from Lua code. The main one is, of course, \tt{torch}, which is installed at the beginning.
Not all the packages are included at first installation, but it is easy to add a new one by the shell command: 
\begin{verbatim}
luarocks install packagename
\end{verbatim}
where \tt{luarocks} is the package manager, and \tt{packagename} is the name of the package you want to install.

\subsubsection*{The \tt{nn} package}
All (almost) you need to create (almost) any kind of ANNs is contained in the \tt{nn} package (which is usually automatically installed).
Every element inside the package inherits from the abstract Lua class \tt{nn.Module}. The main state variables are \tt{output} and \tt{gradInput}, where the result of forward and backward steps (in back-propagation) will be stored.
\tt{forward} and \tt{backward} are methods of such class (which can be accessed by the \emph{object:method()} notation). They invoke \tt{updateOutput} and \tt{updateGradInput} respectively, that here are abstract and the definition must be in the derived classes.

The main advantage of this package is that all the gradients computations in the back-propagation step are automatically realized thanks to these built-in functions. The only requirement is to call the \tt{forward} step before the \tt{backward}.
 
The weights of the network will be updated by the method \tt{updateGradParameters}, assigning a new value to each parameter of a module (according to the Gradient Descent rule) exploiting the \emph{learning rate} passed an argument of the function. 
 
The bricks you can use to construct a network can be divided as follows:
\begin{itemize}
\item \textbf{Simple layers:} the common modules to implement a layer. The main is \tt{nn.Linear}, computing a basic linear transformation.
\item \textbf{Transfer functions:} here you can find many activation functions, such as \tt{nn.Sigmoid} or \tt{nn.Tanh}
\item \textbf{Criterions:} loss functions for supervised tasks, a for instance is \tt{nn.MSECriterion}
\item \textbf{Containers:} abstract modules that allow us to build multi-layered networks. \tt{nn.Sequential} connect several layers in a feed-forward manner. \tt{nn.Parallel} and \tt{nn.Concat} are important to build more complex structure, where the input flows in separated architectures.
Layers, activation functions and even criterions can be added inside those containers.
\end{itemize}

For detailed documentation of the \tt{nn} package we refer to the official webpage\footnote{\url{https://github.com/torch/nn}}. Another useful package for whom could be interested on building more complex architectures can be found at the \tt{nngraph} repository\footnote{Detailed documentation at \url{https://github.com/torch/nngraph}}.

\subsubsection*{CUDA\Cpr computing}

Since C++/Cuda programming and integration are not trivial to develop, it is important to have an interface as simple as possible linking such tools.
Torch provides a clean solution for that with the two dedicated packages \tt{cutorch} and \tt{cunn} (requiring, of course, a capable GPU and CUDA\Cpr installed). All the objects can be transferred into the memory of GPUs by the method \tt{:cuda()} and then back to the CPU by \tt{:double()}. Operations are executed on the hardware of the involved objects and are possible only among variables from the same unit. In \RefList{torchcudasamples} we show some examples of correct and wrong statements. 
\lcode
\begin{lstlisting}[caption= Samples of CUDA operations.,label=torchcudasamples]
local cpuTensor1 = torch.Tensor(3,3):fill(1)
local cpuTensor2 = torch.Tensor(3,3):fill(2)
local cudaTensor1 = torch.Tensor(3,3):fill(3):cuda()
local cudaTensor2 = torch.Tensor(3,3):fill(4):cuda()

cpuTensor1:cmul(cpuTensor2) -- OK
cpuTensor1:cmul(cudaTensor2) -- WRONG
cudaTensor1:cmul(cudaTensor2) -- OK
cudaTensor1:cmul(cpuTensor2) -- WRONG
\end{lstlisting}

\subsection{Setting up the XOR experiment} 
In order to give a concrete feeling about the presented tools, we show some examples on the classical XOR problem as in the previous section. The code showed here below can be found in the Torch section of the document's GitHub repository\footnote{Available at \url{https://github.com/AILabUSiena/NeuralNetworksForBeginners/tree/master/torch/xor}.} and can be useful to play with the parameters and become more familiar with the environment.


\subsubsection*{Architecture} 
When writing a a script, the first command is usually the import of all the necessary packages by the keyword \tt{require}. In this case, only the \tt{nn} toolbox is required:
\begin{lstlisting}
require 'nn'
\end{lstlisting}
We define a standard ANNs with one hidden layer composed by $2$ hidden units, the \emph{hyperbolic tangen} (\tt{tanh}) as transfer function and identity as output function. The structure of the network will be stored in a container where all the necessary modules will be added. A standard feed-forward architecture can be defined into a \tt{Sequential} container, which we named \emph{mlp}. The network can be then assembled by adding sequentially all the desired modules by the function \tt{add()}:

\begin{lstlisting}[caption = Code to create a 2-layer ANNs,  label = torchXorNet ]
mlp = nn.Sequential() -- container initialization
inputs = 2; outputs = 1; HUs = 2; -- general options
mlp:add(nn.Linear(inputs, HUs)) -- first linear layer
mlp:add(nn.Tanh()) -- activation for the hidden layer
mlp:add(nn.Linear(HUs, outputs)) -- output layer
\end{lstlisting}

\subsubsection*{Dataset} 
The training set will be composed by a tensor of $4$ samples (organized again column-wise) paired with a tensor of targets. Usually, \emph{true} and \emph{false} boolean values are respectively associated to $1$ and $0$. However, just to propose an equivalent but different approach, here we shift both values by $-0.5$, so they will be in $[-0.5,0.5]$ as showed in \RefList{torchXorDataset}. 
Both input and target are initialized with \emph{false} values (a tensor filled with 0), and then \emph{true} values are placed according to the XOR truth table.

\lcode
\begin{lstlisting}[caption= creation of $4$ examples and their targets , label = torchXorDataset]
local dataset = torch.Tensor(4,2):fill(0)
local target = torch.Tensor(4,1):fill(0)
dataset[2][1] = 1 -- True False
dataset[3][2] = 1 -- False True
dataset[4][1] = 1; dataset[4][2] = 1 -- True True
dataset = dataset:add(-0.5) -- shift true and false by -0.5
target[2][1] = 1; target[3][1] = 1
target = target:add(-0.5)
\end{lstlisting}

\subsubsection*{Training}
We set up a full--batch mode learning, i.e. we update the parameters after accumulating the gradients over the whole dataset. We exploit the following function:
\begin{description}
\item[\tt{forward(}$input$\tt{)}] returns the output of the multi layer perceptron w.r.t the given input; it updates the input/output states variables of each modules, preparing the network for the backward step; its output will be immediately passed to the loss function to compute the error.
\item[\tt{zeroGradParameters}()] resets to null values the state of the gradients of the all the parameters.
\item[\tt{backward(}$gradients$\tt{)}] actually computes and accumulates (averaging them on the number of samples) the gradients with respect to the weights of the network, given the data in input and the gradient of the loss function.
\item[\tt{updateParameters(}$learningrate$\tt{)}] modifies the weights according to the Gradient Descent procedure using the learning rate as input argument.
\end{description}

\noindent As loss function we use the Mean Square Error, created by the statement:
\lcode
\begin{lstlisting}
criterion = nn.MSECriterion()
\end{lstlisting}
When a criterion is forwarded, it returns the error between the two input arguments. It updates its own modules state variable and gets ready to compute the gradients tensor of the loss in the backward step, which will be back-propagated through the multilayer perceptron. As a \tt{nn} modules, all the possible criterions used the functions \tt{forward()} and \tt{backward()} as the others. The whole training procedure can be set up by:
\begin{lstlisting}[caption=training of the network,label=torchXortrain]
nepochs = 1000; learning_rate = 0.05; -- general options
local loss = torch.Tensor(nepochs):fill(0); -- training loss initialization
for i = 1,nepochs do -- iterating over 1000 epochs
  local input = dataset
  local outputs = mlp:forward(input) -- network forward step
  loss[i] = criterion:forward(outputs, target) -- error evaluation
  mlp:zeroGradParameters()  -- zero reset of gradients
  local gradients = criterion:backward(mlp.output, target) -- loss gradients
  mlp:backward(input, gradients) -- network backward step
  mlp:updateParameters(learning_rate)  -- update parameters given the learning rate
end
\end{lstlisting}

\subsection{Stopping criterions and Regularization}
Since the training procedure is manually defined, particular stopping criterion are completely up to the user. The simplest one, based on the reaching of a fixed number of epochs explicitly depends of the upper bound of the \tt{for} cycle. Since other methods are related to the presence of a validation set, we will define an example of early stopping criterion in \RefList{mnisttorchtrain} in \RefSec{mnistsec}. A simple criterion based on the vanishing of the gradients can be simply set up by exploiting the function \tt{getParameters} defined for the modules of \tt{nn}, which returns all the weights and the gradients of the network in two 1-Dimensional vector:
\begin{lstlisting}
param, grad = mlp:getParameters()
\end{lstlisting}
A simple check on the minimum value of the absolute values of gradients saved in \tt{grad} can be used to stop the training procedure.

Another regularization method can be accomplished by implementing the weight decay method as shown in~\RefList{torchweightdecay}. The presented code is intended to be an introductory example even to understand the class inheritance mechanisms in Lua and Torch.

\lcode
\begin{lstlisting}[caption= Code to implement the weight decay regularization, label=torchweightdecay]
-- defining class inheritances
local WeightDecay, parent = torch.class('nn.WeightDecayWrapper', 'nn.Sequential')

function WeightDecay:__init() -- constructor
	parent.__init(self)
	self.weightDecay = 0 -- self is a keyword referring to the abstract object
	self.currentOutput = 0
end

function WeightDecay:getWeightDecay(alpha)
	local alpha = alpha or 0
	local  weightDecay = 0
	for i=1,#self.modules do
		local params,_ = self.modules[i]:parameters()
		if params then
			for j=1,#params do
				weightDecay = weightDecay + torch.dot(params[j], params[j])*alpha/2
			end
		end
	end
	self.weightDecay = weightDecay
	return self.weightDecay
end
		
function WeightDecay:updateParameters(learningRate,alpha)
	local alpha = alpha or 0
	for i=1,#self.modules do
		local params, gradParams = self.modules[i]:parameters()
		if params then
			for j=1,#params do
				params[j]:add(-learningRate, gradParams[j] + (alpha*params[j]))
			end
		end
	end
end
\end{lstlisting}

\noindent To implement the weight decay coherently with the \tt{nn} package, we need to
create a novel class, inheriting from \tt{nn.Sequential}, that overloads the method \tt{updateParameters()} of the \tt{nn.Module}. We first have to declare a new class name in Torch, where you can optionally specify the parent class. In our case the new class has been called \tt{nn.WeightDecayWrapper}, while the class it inherits from is the Container \tt{nn.Sequential}. The constructor is defined within the function \tt{WeightDecay:\_\_init()}. In the scope of this function the variable \tt{self} is a table used to refer to all the attributes and methods of the abstract object. The calling of the function \tt{\_\_init()} from the \tt{parent} class automatically add all the original properties. The functions \tt{WeigthDecay:getWeigthDecay()} and \tt{WeigthDecay:updateParameters()} compute respectively the weight decay and its gradient. Both methods loop over all the modules of the container (the symbol \# returns the number-indexed element of a table) and, for each one that has parameters, use them in order to compute either the error or the gradients coming from the weight decay contribution. The argument \tt{alpha} represent the regularization parameter of the weight decay and, if not provided, is assumed null. It is also worth to mention the fact that, \tt{WeigthDecay:updateParameters()} overloads the method that implemented in \tt{nn.Module}, updating the parameters according to the standard Gradient Descent rule. At this point, an ANN expecting a possible weight decay regularization can be declared by replacing the \tt{nn.Sequential} container by the proposed \tt{nn.WeightDecayWrapper}.

\subsection{Drawing Separation Surfaces}

\begin{figure}[H]
	\begin{center}
	\begin{tabular}{cc} 		
		\includegraphics[height = 6.5cm,width=7.7cm]{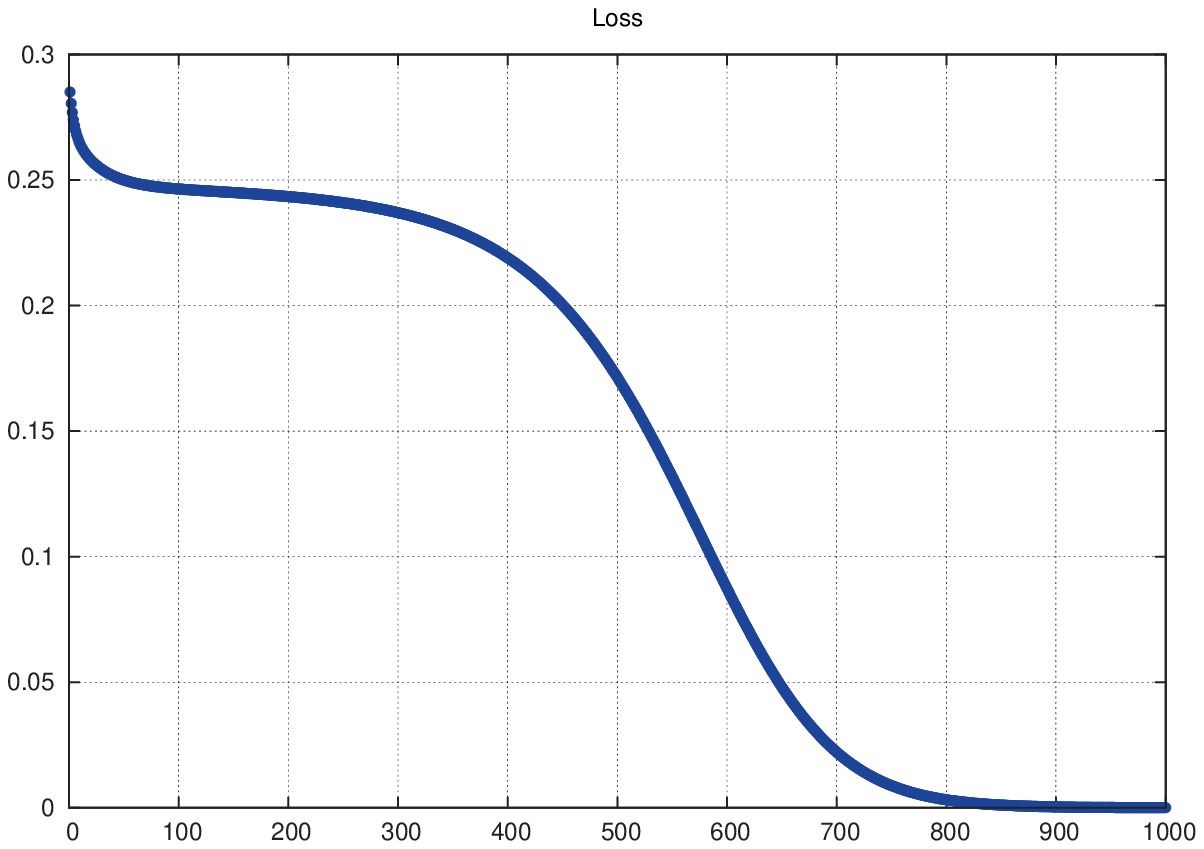}	&
		\includegraphics[height = 6.5cm,width=7.7cm]{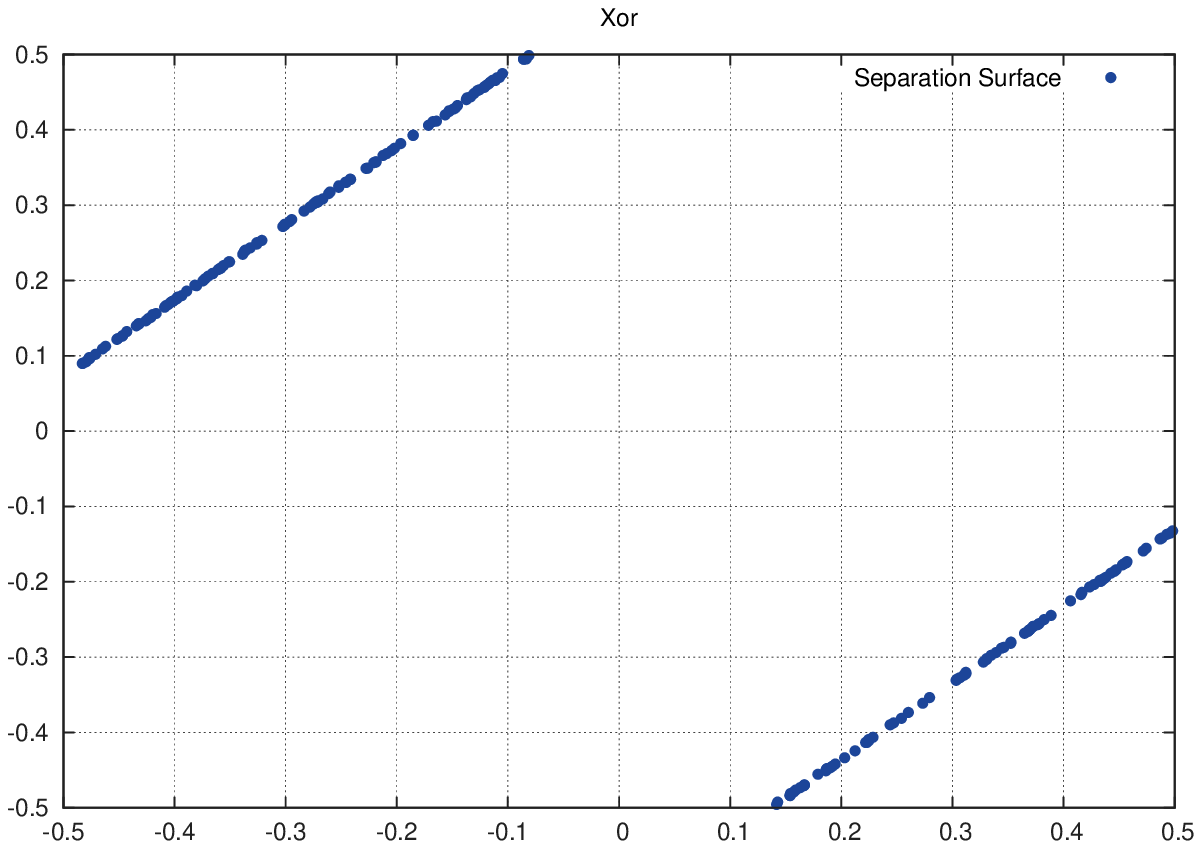}\\
		(a) & (b) 
		\end{tabular}			
	\end{center}
	\caption{(a) Trend of loss versus the number of epochs. (b) The estimated separation surface obtained by a 2-Layers ANN composed by 2 Hidden Units, \emph{Hyperbolic Tangent} as activation and linear output.}\label{torchPlots}
\end{figure}

In this framework, data visualization is allowed by the package \tt{gnuplot}, which provides some tools to plot points, lines, curves and so on. For example, in the training procedure presented in~\RefList{torchXortrain}, a vector storing the penalty evaluated at each epoch is produced. To have an idea of the state of the network during training, we can save an image file containing the trend of the error by the code in~\RefList{torchlossplot}, whom output is shown in~\RefFig{torchPlots}(a).\\

\begin{lstlisting}[caption = Error evaluated at each training epoch., label = torchlossplot]
	gnuplot.epsfigure('XORloss.eps') -- create an eps file with the image
	gnuplot.plot({'loss function', torch.range(1,nepochs),loss}) -- loss trend plot
	gnuplot.title('Loss') 
	gnuplot.grid(true)
	gnuplot.plotflush()
	gnuplot.figure()
\end{lstlisting}
Torch does not have dedicated functions to visualize separation surfaces produced on data and, hence, we generate a random grid across the input space, plotting only those points predicted close enough (with respect to a  a certain threshold) to the half of possible target (0 in this case). The correspondent result, showed in~\RefFig{torchPlots}(b), is generated by the code in~\RefList{torchSepSurplot}, exploiting the fact that Lua support the logical indexing as in Matlab.

\lcode
\begin{lstlisting}[caption =Drawing separation surfaces in torch, label = torchSepSurplot]
local eps = 2e-3; -- separation threshold
local  n = 1000000 -- gridding size
-- input space grid by n 2D-points uniformly distributed in [-0.5,0.5]
local x = torch.rand(n,2):add(-0.5)
local mlpOutput = mlp:forward(x) -- evaluation
-- compare whether the absolute value of the network output is less equal than eps
local mask = torch.le(torch.abs(mlpOutput),mlpOutput:clone():fill(eps)) 
if torch.sum(mask) > 0 then
  gnuplot.epsfigure('Separation surface.eps')
  local x1 = x:narrow(2,1,1) -- reshaping first input dimension for x axis
  local x2 = x:narrow(2,2,1) -- reshaping secind input dimension for y axis
  -- plotting of the collection of points that match the mask
  gnuplot.plot(x1[mask], x2[mask], '+') 
  gnuplot.title('Separation surface')
  gnuplot.grid(true)
  gnuplot.plotflush() -- save the image
  gnuplot.figure()
end
\end{lstlisting} 

\newpage

\section{TensorFlow} 

\subsection{Introduction}
TensorFlow~\cite{abadi2016tensorflow} is an open source software library for numerical computation and is the youngest with respect to the others Machine Learning frameworks. It was originally developed by researchers and engineers from the Google Brain Team, with the purpose of encourage research on deep architectures. Nevertheless, the environment provides a large set of tools suitable for several domains of numerical programming. The computation is conceived under the concept of \emph{Data Flow Graphs}. Nodes in the graph represent mathematical operations, while the graph edges represent tensors (multidimensional data arrays). The core of the package is written in C++, but provides a well documented Python API. The main characteristic is its symbolic approach, which allows a general definition of a forward models, leaving the computation of the correspondent derivatives entirely to the environment itself.

\subsection{Getting started}
	\subsubsection{Python}
	A TensorFlow model can be easily written using Python, a very intuitive object-oriented programming language. Python is distributed with an open-source license for commercial use too. It offers a nice integration with many other programming languages and provides an extended standard library which includes \tt{numpy} (modules designed for matrix operations, very similar to the Matlab syntax). Python runs on Windows, Linux/Unix, Mac OS X and other operative systems.
	
\subsubsection{TensorFlow environment}
Assuming that the reader is familiar with Python, here we present the building blocks of TensorFlow framework:

\paragraph{The Data Flow Graph} To leverage the parallel computational power of multi-core CPU, GPU and even clusters of GPUs, the dynamic of the numerical computations has been conceived as a directed graph, where each node represents a mathematical operation and the edges describe the input/output relation between nodes.
		
\paragraph{Tensor} It is a typed n-dimensional array that flows through the Data Flow Graph.
		
\paragraph{Variable} Symbolic objects designed to represent parameters. They are exploited to compute the derivatives at a symbolical level, but in general must be explicitly initialized in a session.
		
\paragraph{Optimizer} It is the component which provides methods to compute gradients from the loss function and to apply back-propagation through all the variables. A collection is available in TensorFlow to implement classic optimization algorithms.
	
\paragraph{Session}  A graph must be launched in a Session, which places the graph onto CPU or GPU and provides methods to run computation.

	\subsubsection{Installation}
	Information about download and installation of Python and TensorFlow are available in the official webpages\footnote{Python webpage: \url{https://www.python.org/}, TensorFlow webpage: \url{https://www.tensorflow.org/}}. Notice that a dedicated procedure must be followed for GPU installation. It's worth a quick remark on the CUDA\Cpr versions. Indeed,  versions from 7.0 are officially supported, but the installation could be not straightforward in versions preceding the  8.0. Moreover, a registration to the \emph{Accelerate Computing Developer Program}\footnote{\url{https://developer.nvidia.com/accelerated-computing-developer}} is required to install the package \tt{cuDNN}, which is mandatory to enable GPU support.

\subsection{Setting up the XOR experiment}

As in the previous sections of this tutorial, we show how to start managing the TensorFlow framework by facing the simple XOR classification problem by a standard ANN.

\subsubsection*{Import tensor flow} 
At the beginning, as for every Python library, we need to import the TensorFlow package by:
	
	\pcode
	\begin{lstlisting}
import tensorflow as tf
	\end{lstlisting}
ciao
	
\subsubsection*{Dataset definition} 
Again, data can be defined as two matrices containing the input data and its correspondent target, called $X$ and $Y$ respectively. Data can be defined as a list or \tt{numpy} array. After they will be used to fill the placeholder that actually define a type and dimensionality.

	 \begin{lstlisting}
X = [[0,0],[0,1],[1,0],[1,1]]
Y = [[0],[1],[1],[0]]
	 \end{lstlisting}
	 
\subsubsection*{Placeholders} TensorFlow provides Placeholders which are symbolic variables representing data during the computation. A Placeholders object have to be initialized with given type and dimensionality, suitable to represent the desired element. In this case we define two object \tt{x\_} and \tt{y\_} respectively for input data and target:
	 \begin{lstlisting}
x_ = tf.placeholder(tf.float32, shape=[4,2])
y_ = tf.placeholder(tf.float32, shape=[4,1])
	 \end{lstlisting}
	 
\subsubsection*{Model definition} 
The description of the network depends essentially on its architecture and parameters (weights and biases). Since the parameters have to be estimated, they are defined as the variabile of the model, whereas the architecture is determined by the configuration of symbolic operations. For a 2-layers ANN we can define:
\begin{lstlisting}
# Hidden units
HU = 3	 
# 1st layer
W1 = tf.Variable(tf.random_uniform([2,HU], -1.0, 1.0)) # weights matrix
b1 = tf.Variable(tf.zeros([HU]))	 # bias
O = tf.nn.sigmoid(tf.matmul(x_, W1) + b1) # non-linear activation output
# 2nd layer
W2 = tf.Variable(tf.random_uniform([HU,1], -1.0, 1.0))
b2 = tf.Variable(tf.zeros([1]))
y = tf.nn.sigmoid(tf.matmul(O, W2) + b2) 
\end{lstlisting}

\noindent The \tt{matmul()} function performs tensors multiplication. \tt{Variable()} is the constructor of the class variable. It needs an initialization value which must be a tensor. The function \tt{random\_uniform()} returns a tensor of a specified shape, filled with valued picked from a uniform distribution between two specified values. The \tt{nn} module contains the most common activation functions, taking as input a tensor and evaluating the non-linear transferring component-wise (the \emph{Logistic Sigmoid} is chosen in the reported example by \tt{tf.nn.sigmoid()}).
		  
\subsubsection*{Loss and optimizer definition} The cost function and the optimizer are defined by the following two lines
	 
 \begin{lstlisting}
 # quadratic loss function
cost = tf.reduce_sum(tf.square(y_ - y), reduction_indices=[0])
# optimizing the function cost by gradient descent with learning step 0.1
train_step = tf.train.GradientDescentOptimizer(0.1).minimize(cost)
\end{lstlisting}
TensorFlow provides functions to perform common operations between tensors. The function \tt{reduce\_sum()} for example reduces the tensor to one (or more) dimension, by summing up along the specified dimension. The \tt{train} module provide the most common optimizers, which will be employed during the training process. The previous code chose the \emph{Gradient Descent} algorithm to optimize the network parameters, with respect to the penalty function defined in \tt{cost} by using a \emph{learning rate} equal to 0.1.
 
\subsubsection*{Start the session}
At this point the variables are still not initialized. The whole graph exist at a symbolic level, but it is instantiated when creating a session. For example, placeholders are fed with the assigned elements in this moment.
	  
	  \begin{lstlisting}
% Create session
sess = tf.Session()
% Initialize variables
sess.run(tf.initialize_all_variables())
	  \end{lstlisting}
More specifically, \tt{initialize\_all\_variables()} creates an operation (a node in the Data Flow Graph) running variables initializer. The function \tt{Session()} creates an instance of the class \emph{session}, while the correspondent method \tt{run()} moves for the first time the Data Flow Graph on CPU/GPU, allocates variables and fills them with the initial values.

\subsubsection*{Training} 
The training phase can be defined in a \tt{for} loop where each iteration represent a single gradient descend epoch. In the following code, some printing on the training information are added each 100 epochs.
	  \begin{lstlisting}
Epochs = 5000 # Number of iterations
	  
for i in range(Epochs):
	sess.run(train_step, feed_dict={x_: X, y_: Y}) # optimizer step
	if i % 100 == 0:
		print('Epoch ', i)
		print('Cost ', sess.run(cost, feed_dict={x_: X, y_: Y}))
	  \end{lstlisting}
The \tt{sess.run()}  calling runs the operations previously defined for the first argument, which in this case is an optimizer step (defined by \tt{train\_step}). The second (optional) argument for \tt{run} is a dictionary \tt{feed\_dict}, pairing each placeholder with the correspondent input. The function \tt{run()} is used also to evaluate the cost each 100 epochs. 

\subsubsection*{Evaluation} 
The performance of the trained model can be easily evaluated by: 
	  \begin{lstlisting}
correct_prediction = abs(y_ - y) < 0.5 
cast = tf.cast(correct_prediction, "float")
accuracy = tf.reduce_mean(cast)
	  
yy, aa = sess.run([y, accuracy],feed_dict={x_: X, y_: Y})
	  
print "Output: ", yy
print "Accuracy: ", aa
	  \end{lstlisting}
	  
\subsection*{Draw separation surfaces}
\begin{figure}[H]
 \begin{center}
\includegraphics{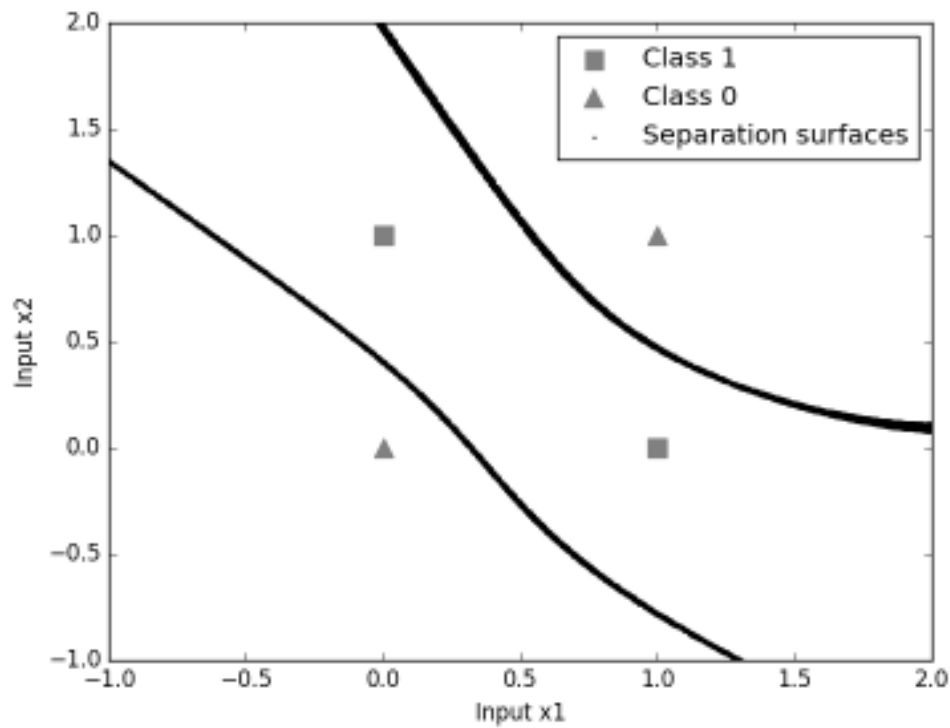}
\end{center}
\caption{Separation surfaces on the XOR classification task obtained by 2-layer ANN with 3 Hidden Units and the \emph{Logistic Sigmoid} as activation and output function.}\label{tfsepsurf}
 \end{figure}

In order to visualize separation surfaces computed by the network, it can be useful to generate a random sample of points on which test results, as showed in~\RefFig{tfsepsurf}.
	\begin{lstlisting}[caption=Draw separation surfaces]
plt.figure()
# Plotting dataset
c1 = plt.scatter([1,0], [0,1], marker='s', color='gray', s=100)
c0 = plt.scatter([1,0], [1,0], marker='^', color='gray', s=100)
# Generating points in [-1,2]x[-1,2]
DATA_x = (np.random.rand(10**6,2)*3)-1
DATA_y = sess.run(y,feed_dict={x_: DATA_x})
# Selecting borderline predictions
ind = np.where(np.logical_and(0.49 < DATA_y, DATA_y< 0.51))[0]
DATA_ind = DATA_x[ind]
# Plotting separation surfaces
ss = plt.scatter(DATA_ind[:,0], DATA_ind[:,1], marker='_', color='black', s=5)
# Some figure's settings
plt.legend((c1, c0, ss), ('Class 1', 'Class 0', 'Separation surfaces'), scatterpoints=1)
plt.xlabel('Input x1')
plt.ylabel('Input x2')
plt.axis([-1,2,-1,2])
plt.show()
	\end{lstlisting}
	  
\newpage

\section{MNIST Handwritten Characters Recognition}\label{mnistsec}

\begin{figure}[H]
 \begin{center}
 \includegraphics{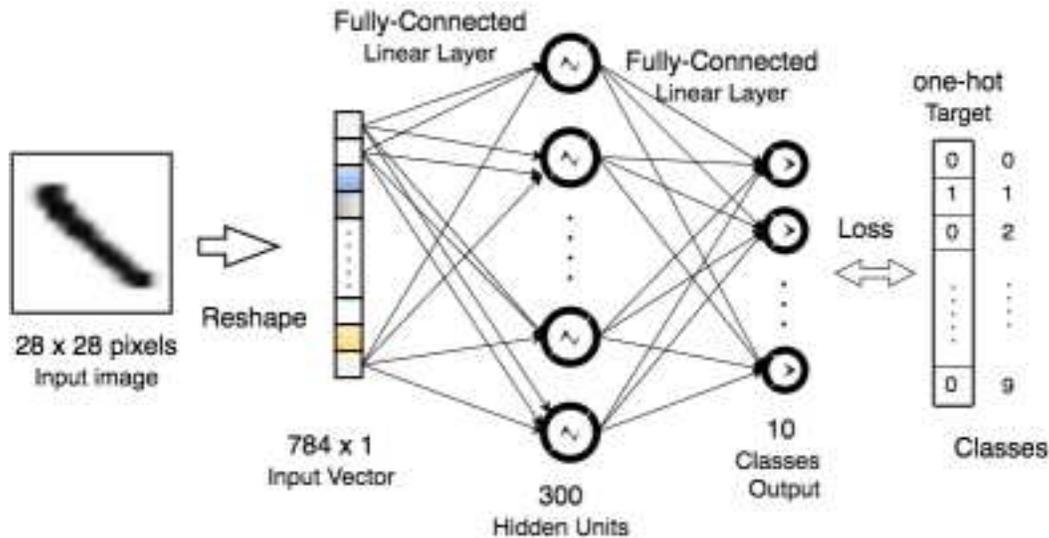}
\end{center}
\caption{General architecture of a 2-Layer network model proposed to face the MNIST data.}\label{mnistmod}
 \end{figure}

In this Section we show how to set up a 2-Layer ANN in order to face the MNIST \cite{lecunn98} classification problem, a well known data set for handwritten characters recognition. It is extensively used to test and compare general Machine Learning algorithms and Computer Vision methods. Data are provided as $28 \times 28$ pixels (grayscale) images of handwritten digits. The training and test sets contain respectively 60,000 and 10,000 instances. Files .zip are available at the official site\footnote{\url{http://yann.lecun.com/exdb/mnist/}.}, together with a list of performance achieved by most common algorithms. We show the setting up of a standard 2-Layer ANN with 300 units in the hidden layer, represented in \RefFig{mnistmod}, since it is one of the architecture reported in the official website and the obtained results can be easily compared. The input will be reshaped so as to feed the network with a 1-Dimensional vector with $28\cdot 28 = 784$ elements. Each image is originally represented by a matrix containing the grayscale value of the pixels in $[0,255]$, which will be normalized in $[0,1]$. The output will be a 10 elements prediction vector, since labels for each element will be expressed by the one-hot encoding binary vector of 10 null bits, with only a 1 in the position indicating the class. Activation and penalty functions are different within different environments to provide an overview on different approaches.

\subsection{MNIST on \MTL} \label{matmnist}

 Once data have been downloaded from the official MNIST site, we can use \MTL functions available at the Stanford University site\footnote{\url{http://ufldl.stanford.edu/wiki/index.php/Using\_the\_MNIST\_Dataset}.}  to extract data from files and organize them in \emph{inputSize} --by--\emph{numberOfSamples} matrix form. The extraction routines reshape (so as that each digit is represented by a 1-D column vector of size 784) and normalizes data (so as that each feature lies in the interval $[0,\,1]$) . Once unzipped data and functions in the same folder, it is straightforward to upload images in the \MTL workspace by the \tt{loadMNISTImages} function:

\mcode
\begin{lstlisting}
 images_Train = loadMNISTImages('train-images.idx3-ubyte');
 images_Test = loadMNISTImages('t10k-images.idx3-ubyte');
 images = [ images_Train, images_Test];
\end{lstlisting}


\noindent where training and test set have been grouped in the same matrix to evaluate performance on the provided test set during the training. Correspondent labels can be loaded and grouped in a similar way by the function \tt{loadMNISTLabels}:

\begin{lstlisting}
 labels_Train = loadMNISTLabels('train-labels.idx1-ubyte');
 labels_Test = loadMNISTLabels('t10k-labels.idx1-ubyte');
 labels = [ labels_Train; labels_Test];
\end{lstlisting}


\noindent The original labels are provided as a 1-Dimensional vector containing a number from 0 to 9 according to the correspondent digit. The one-hot encoding target matrix for the whole dataset can be generated exploiting the \MTL function \tt{ind2vec}\footnote{The function \tt{full} prevent for \MTL automatically convert to sparse matrix, which in our tests may cause some problems at the calling of the function \tt{train}.}:

\begin{lstlisting}
labels = full(  ind2vec( labels' + 1 ) );
\end{lstlisting}


\noindent To check the obtained results, we replied one of the 2-layer architectures listed at the official website, which is supposed to reach around 96\% of accuracy with 300 hidden units and  can be initialized by: 

\begin{lstlisting}
nn = patternnet( 300 );
\end{lstlisting}


\noindent As already said, this command creates a 2-Layer ANN where the hidden layer has 300 units and the \emph{Hyperbolic Tangent} as activation, whereas the output function is computed by  the \emph{softmax}. The (default) penalty function is the \emph{Cross-Entropy Criterion}. 

In this case we change the data splitting so as that data used for test comes only from the original test set (which has been concatenated with the training one), prevent to mix samples among Train, Validation and Test set. This step is completely customizable by the method \tt{divideFcn} and the fields of the options \tt{divideParam}. The \tt{divideind} method picks data according to the provided indexes for the data matrix:

\begin{lstlisting}
nn.divideFcn = 'divideind';
nn.divideParam.trainInd = 1:45000;
nn.divideParam.valInd = 45000:60000;
nn.divideParam.testInd = 60000:70000;
\end{lstlisting}


\noindent In this case, we arbitrarily decided to use the last 25\% of the Training data for Validation, since the samples are more or less equally distributed by classes. 

\begin{figure}[H]
 \begin{adjustbox}{max size={\textwidth}}{}
 \def\tsc{0.15}
 \hspace{-0.5cm}
 \begin{tabular}{cc}
\begin{minipage}[t]{0.4\textwidth}
\includegraphics[scale=0.4]{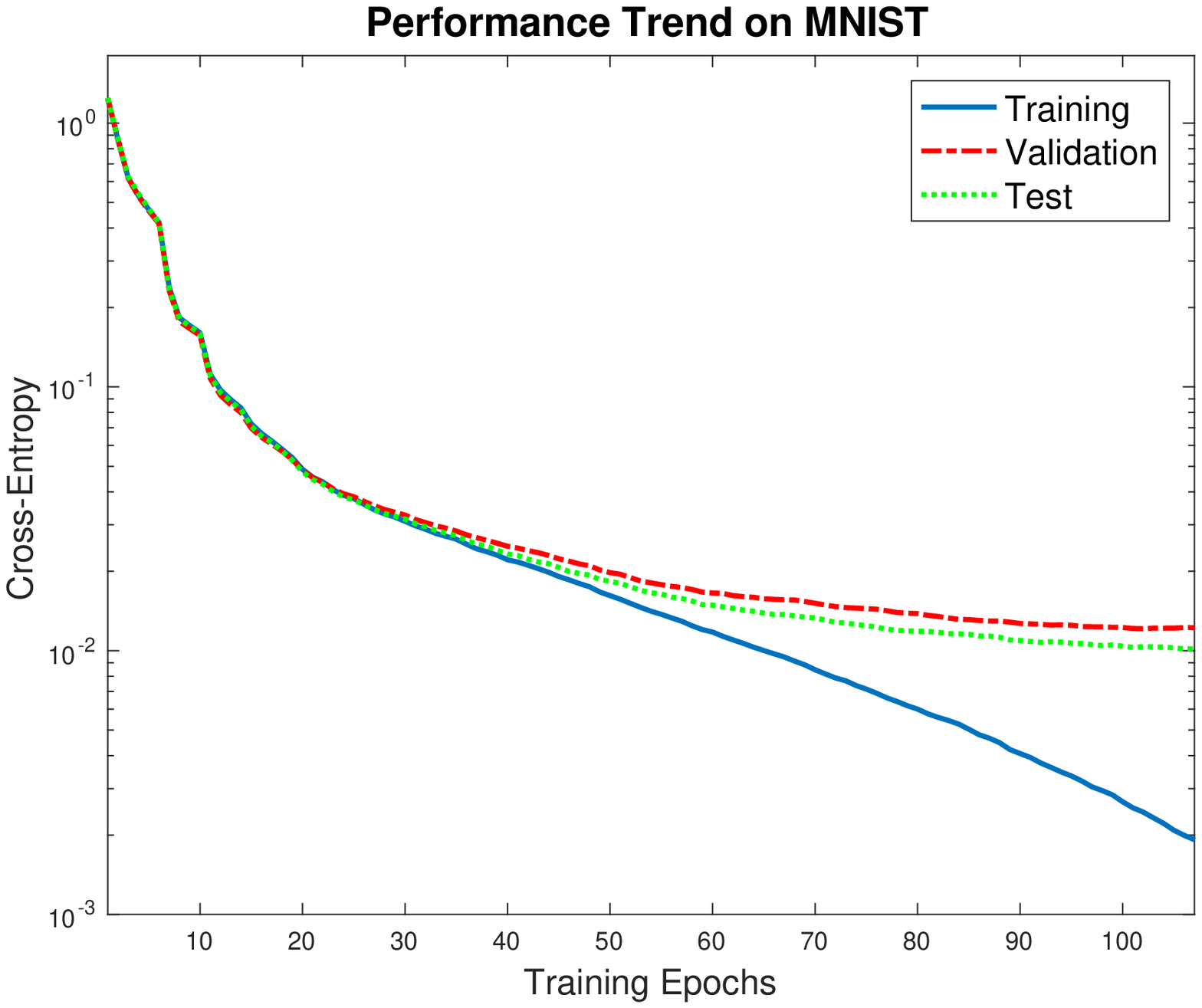} 
\end{minipage} &
\hspace{0.03\textwidth}
\begin{minipage}[c]{0.35\textwidth}
\vspace{-6cm}
\begin{tabular}{c@{\hskip -12pt}c@{\hskip -12pt}c}
\includegraphics[scale=\tsc]{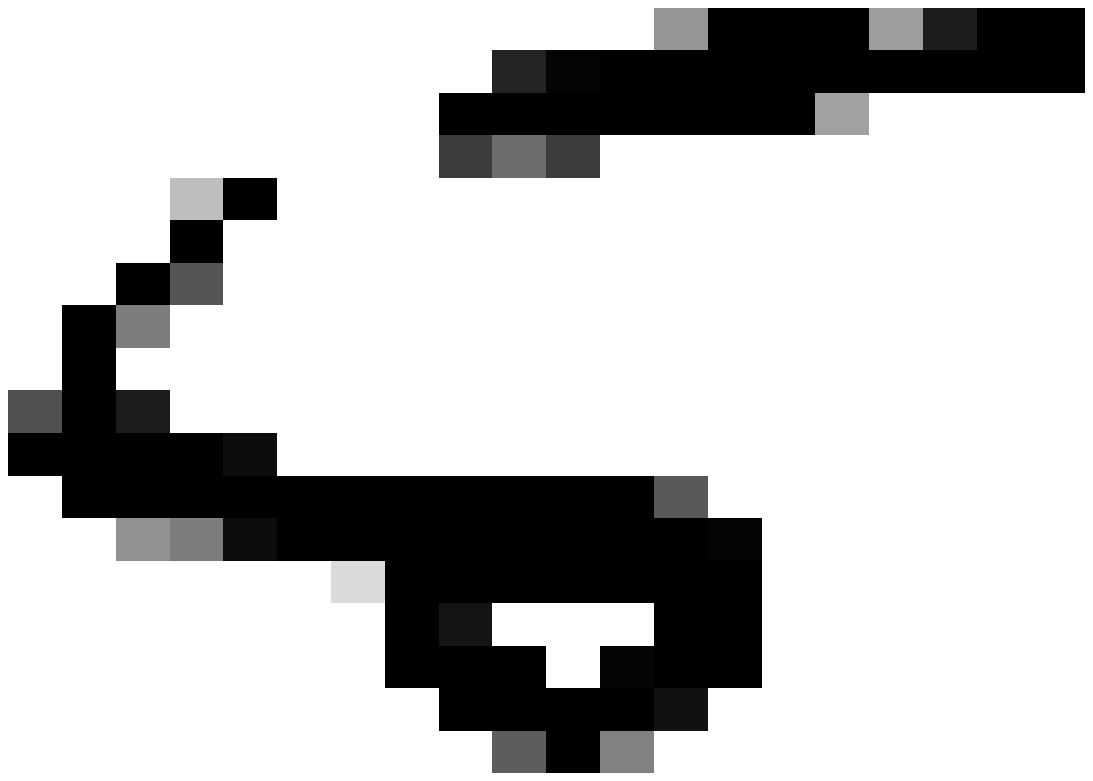}
&
\includegraphics[scale=\tsc]{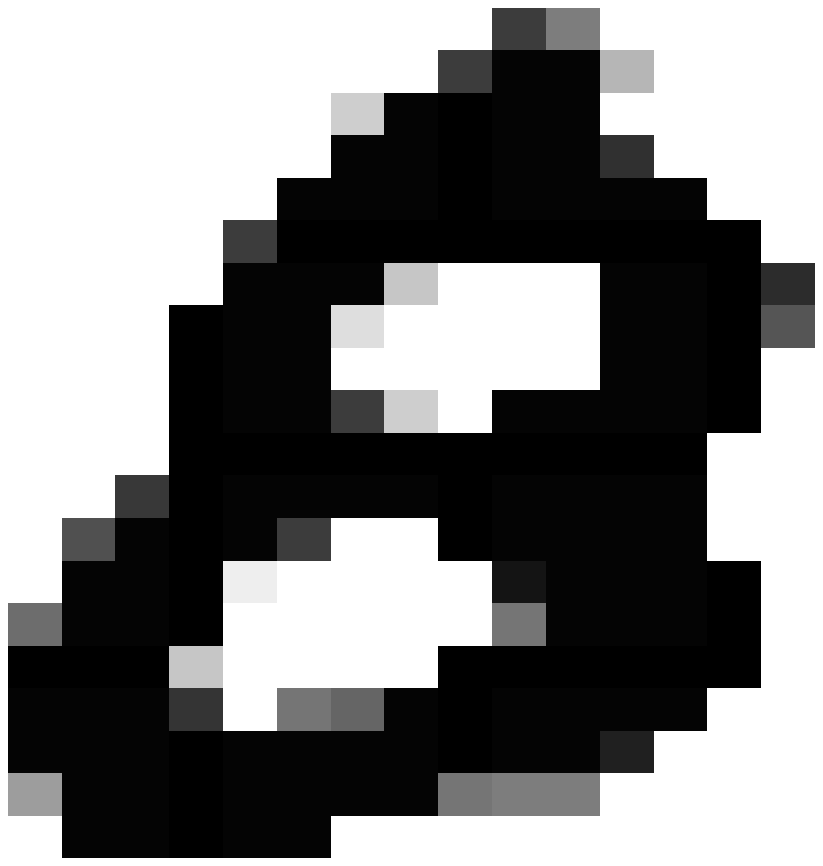}
&
\includegraphics[scale=\tsc]{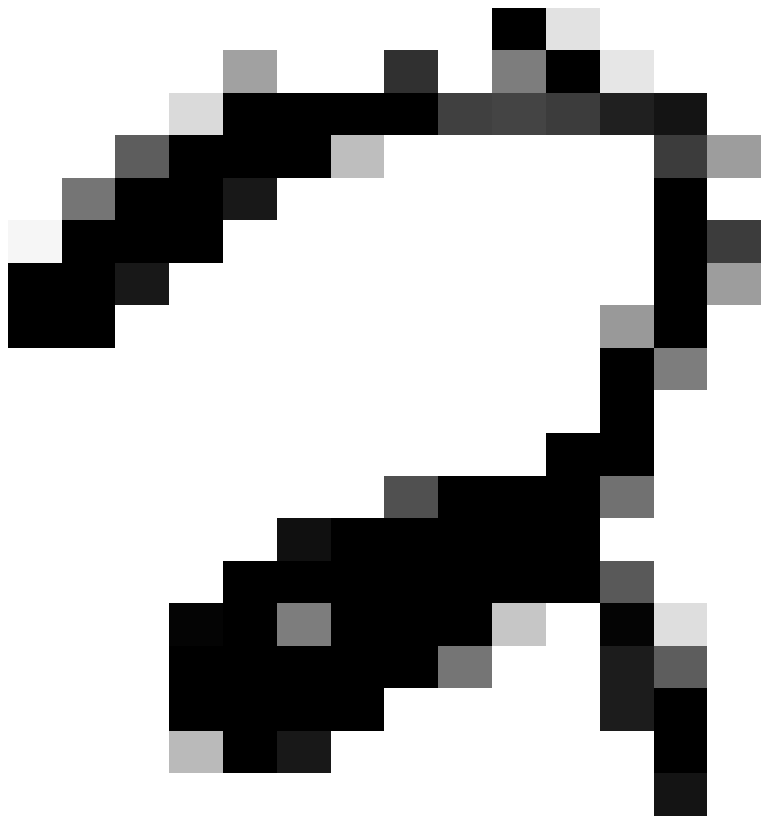}\vspace{-0.5cm} \\
 {\scriptsize $5\rightarrow 6$} & {\scriptsize $8\rightarrow 0$} & {\scriptsize $2\rightarrow 8$} \\
\includegraphics[scale=\tsc]{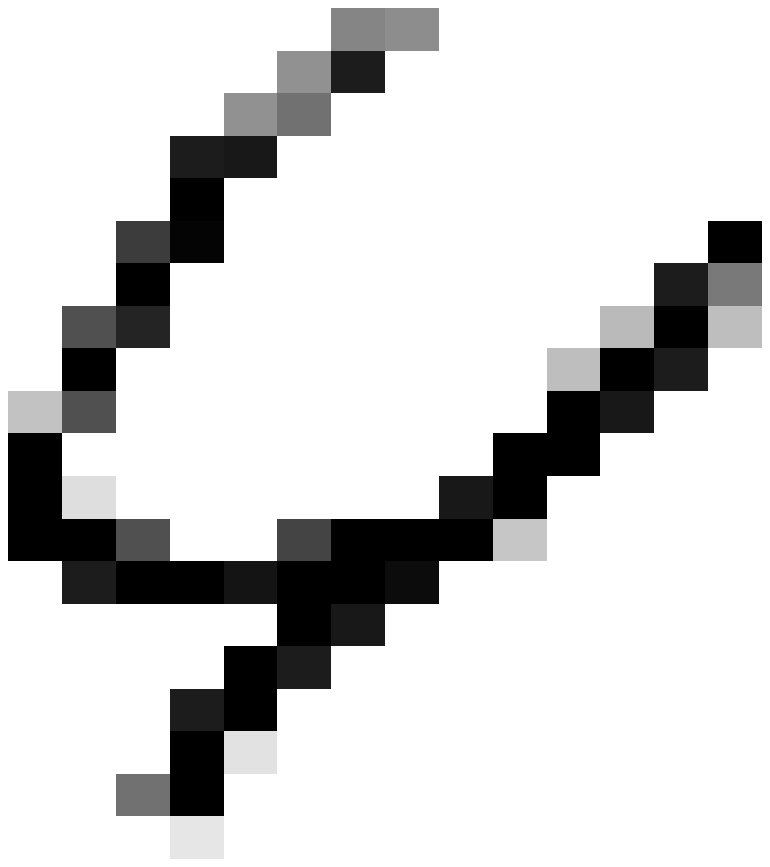} 
&
\includegraphics[scale=\tsc]{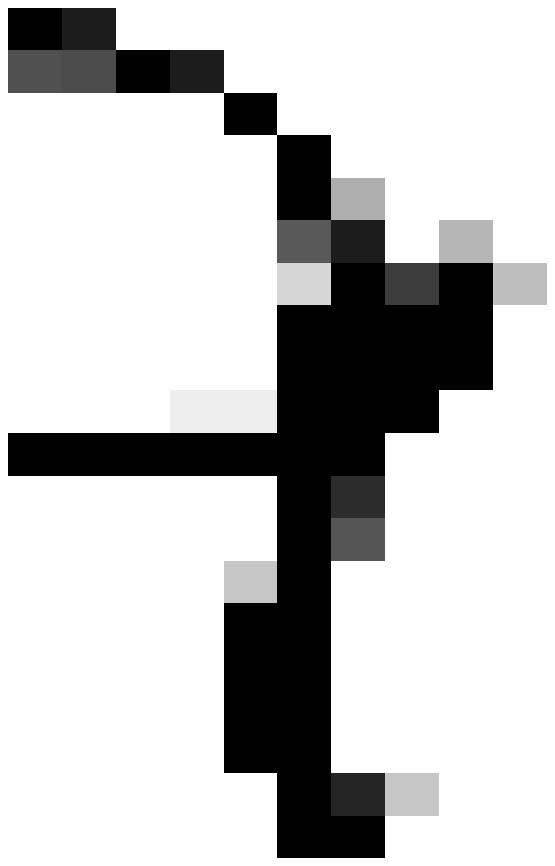} 
&
\includegraphics[scale=\tsc]{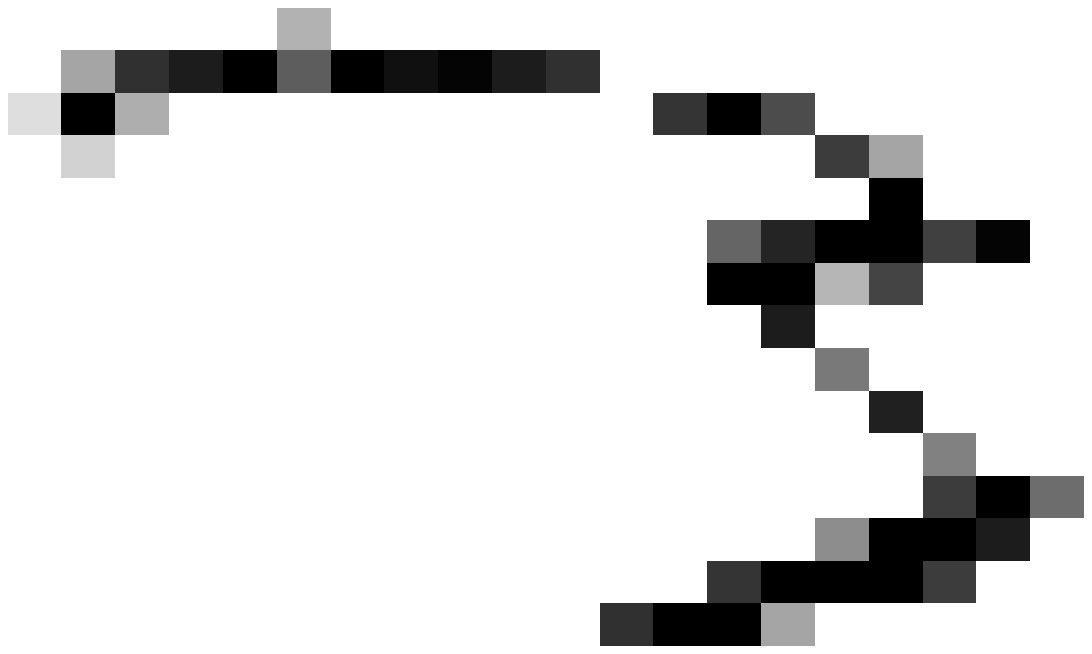}\vspace{-0.5cm} \\
{\scriptsize $4\rightarrow 6$} & {\scriptsize $7\rightarrow 3$} &{\scriptsize $3\rightarrow 7$} \\ 
\end{tabular}
\end{minipage}\\
(a) &  \hspace{3cm}(b)
\end{tabular}
\end{adjustbox}
\caption{On the left the performance on the MNIST dataset during the training of a 2-layer ANN with 300 hidden units. Training is stopped after 107 epochs for validation checking. On the right we report some misclassified samples. The network reaches about 96\% of classification accuracy on the test set (in accordance with the ones provided at the MNIST web page).}\label{mnisttrain}
 \end{figure}

As already said, network training can be started by:

\begin{lstlisting}
[ nn, tr ] = train( nn, images, labels );
\end{lstlisting}

\noindent In the reported case, the training stopped after 107 epochs because of an increasing in the validation error (see \RefSec{regmatlab}). The performance during training are shown in \RefFig{mnisttrain}(a), which is obtained by the following code:
\begin{lstlisting}
% % % % Plotting MNIST training performance

plot(tr.perf,'LineWidth',2); % plot training error
hold on;
plot(tr.vperf,'r-.','LineWidth',2); % plot validation error
plot(tr.tperf,'g:','LineWidth',2); % plot test error
set(gca,'yscale','log'); % setting log scale
axis([1,107,0.001,1.8]);
xlabel('Training Epochs','FontSize',14);
ylabel('Cross-Entropy','FontSize',14);
title('Performance Trend on MNIST','FontSize',16);
h = legend({'Training','Validation','Test'},'Location','NorthEast');
set(h ,'FontSize',14);
\end{lstlisting}

\noindent In \RefFig{mnisttrain}(b) we show some misclassified digits, indicating the original label and the predicted one. The visualization is obtained by the Matlab function \tt{image} (after a reshaping to the original square dimensions and grayscale, multiplying by 255). In \RefList{mnistmatlabperf} we show how to evaluate classification accuracy and confusion matrix on data, which should give coherent results with respect to which reported in the official site for the same architecture (about 4\% error on test set).

\begin{lstlisting}[caption=Performance evaluation of the network trained on the MNIST,label=mnistmatlabperf]
% network evaluation
f = nn( images( :, 1:45000 ) ); % training predictions
fv = nn( images( :, 45001:60000 ) ); % validation predictions
ft = nn( images( :, 60001:end ) ); % test predictions
% classification accuracy
A = mean( vec2ind(f) == vec2ind( labels( :, 1:45000 ) ) );
Av = mean( vec2ind(fv) == vec2ind( labels( :, 45001:60000 ) ) );
At = mean( vec2ind(ft) == vec2ind( labels( :, 60001:end ) ) );
% confusion matrix
C = confusionmat(  vec2ind(f), vec2ind( labels( :, 1:45000 ) ) ) ;
Cv = confusionmat( vec2ind(fv), vec2ind( labels( :, 45001:60000 ) ) );
Ct = confusionmat( vec2ind(ft), vec2ind( labels( :, 60001:end ) ) );
\end{lstlisting}

\subsection{MNIST on Torch} \label{torchmnist}

\lcode

As already said, the Torch environment provides a lot of tools for Machine Learning, included a lot of routines to download and prepare most common Datasets. A wide overview on most useful tutorials, demos and introduction to most common methods can be found in a dedicate webpage\footnote{\url{https://github.com/torch/torch7/wiki/Cheatsheet\#machine-learning}}, including a \emph{Loading popular datasets} section. Here, a link to the MNIST loader page \footnote{\url{https://github.com/andresy/mnist}} is available, where data and all the informations for the correspondent \tt{mnist} package installation are provided. After the installation, data can be loaded by the code in \RefList{mnisttorchdata}.

\begin{lstlisting}[caption=Loading MNIST data,label = mnisttorchdata]

% loading data
images_Train = loadMNISTImages('train-images-idx3-ubyte');
images_Test = loadMNISTImages('t10k-images-idx3-ubyte');
images = [ images_Train, images_Test];

% loading targets
labels_Train = loadMNISTLabels('train-labels-idx1-ubyte');
labels_Test = loadMNISTLabels('t10k-labels-idx1-ubyte');
labels = [ labels_Train; labels_Test];
labels = full(  ind2vec( labels' + 1 ) );
%%
% initialization
nn = patternnet( 300 );
nn.divideFcn = 'divideind';
nn.divideParam.trainInd = 1 : 45000 ;
nn.divideParam.valInd = 45001 : 60000 ;
nn.divideParam.testInd = 60001 : 70000 ;
\end{lstlisting}

\noindent Data will be loaded as a table named \emph{train}, where digits are expressed as a \emph{numberOfSamples}-by-28-by-28 tensor of type \tt{ByteTensor} stored in the field \emph{data}, expressing the value of the gray levels of each pixel between 0 and 255. Targets will be stored as a 1-D vector, expressing the digits labels, in the field \emph{label}. We have to convert data to the \tt{DoubleTensor} format and, again, normalize the input features to have values in $[0,1]$ and reshape the original 3-D tensor in a 1-D input vector.  Labels have to be incremented by 1, since \tt{CrossEntropyCriterion} accepts target indicating the class avoiding null values (i.e. 1 means a sample to belong to the first class and so on). In the last row we perform a random shuffling of data in order to prepare the train/validation splitting by the function:

\begin{lstlisting}
function rndShuffle(dataset)
	-- random data shuffle
	local n = dataset.data:size(1) 
	local perm = torch.LongTensor():randperm(n)
	dataset.data = dataset.data:index(1, perm)
	dataset.label = dataset.label:index(1, perm)
	return dataset
end
\end{lstlisting}

\noindent At this point we can create a \emph{validation} set from the last quarter of the training data by:

\begin{lstlisting}
-- splitting function definition
function splitData(data, p)
	-- splits data depending on the rate p 
	local p = p > 0 and p <= 1 and p or 0.9	
	local trainSize = torch.floor(p*data:size(1))
	return data:narrow(1,1,trainSize),data:narrow(1,1 + trainSize, data:size(1) - trainSize)
end

validation = {}
trainRate = 0.75
train.data, validation.data = splitData(train.data, trainRate)
train.label, validation.label = splitData(train.label, trainRate)
\end{lstlisting}

\noindent The code to build the proposed 2-layer ANN model is reported in \RefList{mnisttorchnet}, where the network is assembled in the \tt{Sequential} container, using this time the \emph{ReLU} as activation for the hidden layer, whereas output and penalty functions are the same used in the previous section (\emph{softmax} and \emph{Cross-Entropy} respectively).

\begin{lstlisting}[caption=Network definition for MNIST data, label=mnisttorchnet]
-- mlpwork definition
require 'nn'

local mlp = nn.Sequential()
mlp:add(nn.Linear(784, 300))         
mlp:add(nn.ReLU()) 
mlp:add(nn.Linear(300, 10))           
mlp:add(nn.SoftMax())                 

-- loss function
local c = nn.CrossEntropyCriterion()
\end{lstlisting}

\noindent The network training can be defined in a way similar to the one proposed in \RefList{torchXortrain}. Because of the width of the training data, this time is more convenient to set up a \emph{minibatch} training as showed in \RefList{mnisttorchtrain}. Moreover, we also define an early stopping criterion which stops the training when the penalty on the validation set start to increase, preventing overfitting problems. The training function expects as inputs the network (named \emph{mlp}), the criterion to evaluate the loss (named \emph{criterion}), training and validation data (named \emph{trainset} and \emph{validation} respectively) organized as a table with fields \emph{data} and \emph{label} as defined in \RefList{mnisttorchdata}. An optional configuration table \emph{options} can be provided, indicating the number of training epochs (\emph{nepochs}), the learning rate (\emph{learning\_rate}), the mini-batch size (\emph{batchSize}) and the number of consecutive increasings in the validation loss which causes a preventive training stop (\emph{maxSteps}). It is worth a remark on the function \tt{split}, defined for the \tt{Tensor} class, used to divide data in batches stored in an indexed table. At the end of the training, a vector containing the loss evaluated at each epoch  is returned. The validation loss is computed with the help of the function \tt{evaluate}, which splits again the computation in smaller batches, preventing from too heavy computations when the number of parameters and samples is very large.

\begin{lstlisting}[caption=Sample of a mini-batch training function, label=mnisttorchtrain]
function training(mlp, criterion, trainset, validation, options)
-- minibatch training
assert(mlp and trainset, "At least 2 arguments are expected")
local nepochs = options and options.nepochs or 1000 -- max number of epoch
local learning_rate = options and options.learning_rate or 0.01 
local batchSize = options and options.batchSize or 32
local maxSteps = options and options.maxSteps or 10 -- max validation fails
local input, target = trainset.data, trainset.label
-- vector for saving loss during epoch
local loss = torch.Tensor(nepochs):typeAs(input):fill(0)
local valLoss = torch.Tensor(nepochs):typeAs(input):fill(0)
-- minibatches splitting 
local minibatches, minitarget = input:split(batchSize), target:split(batchSize)
-- last loss and validation fails for early stopping criterion 
-- based on validation loss
local lastLoss,step = 0,-1
local epoch = 1
while epoch < nepochs and step < maxSteps do
for bi, batch in pairs(minibatches) do
loss[epoch] = loss[epoch] + criterion:forward(mlp:forward(batch), minitarget[bi])
-- reset gradients to null values
mlp:zeroGradParameters()
-- accumulate gradients
mlp:backward(batch, criterion:backward(mlp.output, minitarget[bi]))
-- update parameters
mlp:updateParameters(learning_rate)
if bi%100 == 0 then collectgarbage() end -- cleaning nil variable
end
-- evaluating validation loss
local currLoss = evaluate(mlp, criterion, validation)
valLoss[epoch] = currLoss
if currLoss >= lastLoss*0.9999 then
step = step + 1
else
step = 0
end
lastLoss = currLoss		
xlua.progress(epoch, nepochs) -- printing epochs progress
epoch = epoch + 1
end
if step == maxStep then 
print('Training stopped at Epoch: '..epoch..
' because of Validation Error increasing in the last '
..maxStep..' epochs') 
end
return loss:narrow(1,1,epoch-1)/#minibatches, valLoss:narrow(1,1,epoch-1) 
end

function evaluate(mlp, criterion, dataset, options)
-- evaluate the loss from criterion between mpl predictions
-- by accumulating within minibatches from dataset
assert(mlp and dataset, "At least 2 arguments are expected")
local batchSize = options and options.batchSize or 32
local input, target = dataset.data, dataset.label
local loss = 0
local minibatches, minitarget = input:split(batchSize), target:split(batchSize)
for bi, batch in pairs(minibatches) do
loss = loss + criterion:forward(mlp:forward(batch), minitarget[bi])
end
mlp:zeroGradParameters()
return loss/#minibatches
end
\end{lstlisting}

\noindent  In \RefList{mnisttorchacc} we show how to compute the Confusion Matrix and the Classification Accuracy on data by the function \tt{confusionMtx}, taking in input the network (\emph{mlp}), data (\emph{dataset}) and the expected number of classes (\emph{nclasses}).

\begin{lstlisting}[caption=Evaluating Confusion Matrix and Accuracy, label=mnisttorchacc]
function confusionMtx(mlp, dataset, nclasses)
local input, target = dataset.data, dataset.label
local confMtx = torch.zeros(nclasses,nclasses):typeAs(input)

local prediction = mlp:forward(input)
-- get the position of the unit with the maximum value
local _, pos = torch.max(prediction,2) 
local acc = 0
for i = 1,nclasses do
local predicted = torch.eq(pos,i):typeAs(input)
for j = 1,nclasses do
local truth =  torch.eq(target, j):typeAs(input)			
confMtx[i][j] = torch.cmul(predicted, truth):sum(1)
if i == j then acc = acc + confMtx[i][j] end
end
end	 			
return confMtx, acc/input:size(1)
end	
\end{lstlisting}

\begin{figure}[H]
  \centering
 \includegraphics[height=8cm,width=10cm]{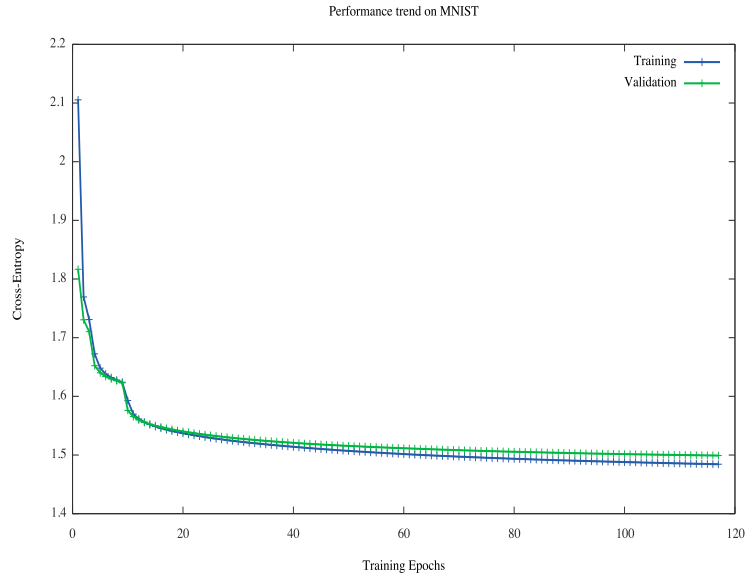} 
 \caption{Confusion matrix on the MNIST test set.}\label{torchmnistcfm}
 \end{figure}

\noindent At this point we can start a trial by the following code:
\begin{lstlisting}
options = {nepochs = 250, batchSize = 64, learning_rate = 0.05, maxStep = 50}
training(mlp, c, train, validation, options)
confMtx, acc = confusionMtx(mlp, test, 10)
\end{lstlisting}

\noindent In this case the training is stopped by the validation criterion after epoch 117, producing a Classification Accuracy on test of about 97\%. In  \RefFig{torchmnistcfm}(a) we report the trend of the error during training. Since in general can be useful to visualize the confusion matrix (which in this case is almost diagonal),  in \RefFig{torchmnistcfm}(b) we show the one obtained by the function \tt{imagesc} from the package \tt{gnuplot}, which just give a color map of the matrix passed as input.

 \subsection{MNIST on Tensor Flow} \label{tfmnist}
Even TensorFlow environment makes available many tutorials and preloaded dataset, including the MNIST. A very fast introductive section for beginners can be found at the official web page\footnote{\url{https://www.tensorflow.org/versions/r0.12/tutorials/mnist/beginners/index.html}}. However, we will show some functions to download and import the dataset in a very simple way. Indeed, data can be directly loaded by:

\pcode
\begin{lstlisting}
from tensorflow.examples.tutorials.mnist import input_data
mnist = input_data.read_data_sets("MNIST_data/", one_hot=True)
\end{lstlisting}

\noindent We firstly define two auxiliary functions. The first (\tt{init\_weights}) will be used to initialize the parameters of the model, whereas the second (\tt{mlp\_output}) to compute the predictions of the model. 

\begin{lstlisting}
def init_weights(shape):
return tf.Variable(tf.random_uniform(shape, -0.1, 0.1))

def mlp_output(X, W_h, W_o, b_h, b_o):

a1 = tf.matmul(X, W_h) + b_h
o1 = tf.nn.relu(a1)  #output layer1

a2 = tf.matmul(o1, W_o) + b_o
o2 = tf.nn.softmax(a2)  #output layer2

return o2
\end{lstlisting}

\noindent Now, with the help of the proposed function \tt{init\_weights}, we define the parameters to be learned during the computation. $W1$ and $b1$ represent respectively the weights and the biases of the hidden layer. Similarly, $W2$ and $b2$ are respectively the weights and the biases of the output layer.

\begin{lstlisting}
W1 = init_weights([x_dim, h_layer_dim])
b1 = init_weights([h_layer_dim])
W2 = init_weights([h_layer_dim, y_dim])
b2 = init_weights([y_dim])
\end{lstlisting}

\noindent Once we defined the weights, we can symbolically compose our model by the calling of our function \tt{mlp\_output}. As in the XOR case, we have to define a placeholder storing the input samples.

\begin{lstlisting}
# Input
x = tf.placeholder(tf.float32, [None, x_dim])
# Prediction
y = mlp_output(x, W1, W2, b1, b2)
\end{lstlisting}

\noindent Then we need to define a cost function and an optimizer. However, this time we add the square of the Euclidean Norm as regularizer, and the global cost function is composed by the \emph{Cross-Entropy} plus the regularization term scaled by a coefficient of $10^{-4}$. At the beginning of the session, TensorFlow moves the Data Flow Graph to the CPUs or GPUs and initializes all variables. 
\pcode
\begin{lstlisting}
# Model Specifications
LEARNING_RATE = 0.5
EPOCHS = 5000
MINI_BATCH_SIZE = 50
# Symbolic variable for the target
y_ = tf.placeholder(tf.float32, [None, y_dim])
# Loss function and optimizer
cross_entropy = tf.reduce_mean(tf.nn.softmax_cross_entropy_with_logits(y, y_))
regularization = (tf.reduce_sum(tf.square(W1), [0, 1]) 
									+ tf.reduce_sum(tf.square(W2), [0, 1]) )
cost = cross_entropy + 10**-4 * regularization
train_step = tf.train.GradientDescentOptimizer(LEARNING_RATE).minimize(cost)
# Start session and initialization
sess = tf.Session()
sess.run(tf.initialize_all_variables())
\end{lstlisting}
TensorFlow provides the function \tt{next\_batch} for the \tt{mnist} class to randomly extract batches of a specified size. Data are split in shuffled partitions of the indicated size and, by means of an implicit counter, the function slide along batches at each calling allowing a fast implementation for mini-batch Gradient Descent method. In our case, we used a \tt{for} loop to scan across batches, executing a training step on the extracted data at each iteration. Once the whole Training set has been processed, the loss on Training, Validation and Test sets is computed. These operations are repeated in a \tt{while} loop, whose steps representing the epochs of training. The loop stops when the maximum number of epochs is reached or the network start to overfit Training data. The early stopping is implemented by checking the Validation error and training is stopped when no improvements are obtained for a fixed number of consecutive epochs (\tt{val\_max\_steps}). The maximum number of epochs and learning rate must be set in advance.

\begin{lstlisting}
# Save values to be plotted
errors_train=[]
errors_test=[]
errors_val=[]

# Early stopping (init variables)
prec_err = 10**6 # just a very big value
val_count = 0
val_max_steps = 5

# Training
BATCH_SIZE = np.shape(mnist.train.images)[0]
MINI_BATCH_SIZE = 50

i = 1
while i <= epochs and val_count < val_max_steps:
	
	# Train over the full batch is performed with mini-batches algorithm
	for j in range(BATCH_SIZE/MINI_BATCH_SIZE):
		batch_xs, batch_ys = mnist.train.next_batch(50)
		sess.run(train_step, feed_dict={x: batch_xs, y_: batch_ys})
		
	# Compute error on validation set and control for early-stopping
	curr_err = sess.run(cross_entropy, 
				feed_dict={x: mnist.validation.images, y_: mnist.validation.labels})
	if curr_err >= prec_err*0.9999:
		val_count = val_count + 1
	else:
		val_count = 0
		
	prec_err = curr_err
	
	# Save values for plot
	errors_val.append(curr_err)
	c_test = sess.run(cross_entropy, 
				feed_dict={x: mnist.test.images, y_: mnist.test.labels})
	errors_test.append(c_test)
	c_train = sess.run(cross_entropy, 
				feed_dict={x: mnist.train.images, y_: mnist.train.labels})
	errors_train.append(c_train)
	
	# Print info about the current epoch
	print "\n\nEPOCH: ",i, "/", epochs
	print "  TRAIN ERR: ", c_train
	print "  VALIDATION ERR: ", curr_err
	print "  TEST ERR: ", c_test
	print "\n(Early stopping criterion: ", val_count, "/", val_max_steps, ")"
	
	# Increment epochs-index
	i = i+1
\end{lstlisting}

\noindent The prediction accuracy of the trained model can be evaluated over the test set in way similar to the one presented for the XOR problem. This time we need to exploit the \emph{argmax} function to convert the one-hot encoding in the correspondent labels.

\begin{lstlisting}
# Symbolic formulas 
correct_prediction = tf.equal(tf.argmax(y,1), tf.argmax(y_,1))
accuracy = tf.reduce_mean(tf.cast(correct_prediction, tf.float32))
# Compute accuracy on the test set
aa = sess.run(accuracy, feed_dict={x: mnist.test.images, y_: mnist.test.labels})
print "Accuracy: ", aa
\end{lstlisting}

\begin{figure}[H]
 \begin{center}
\includegraphics{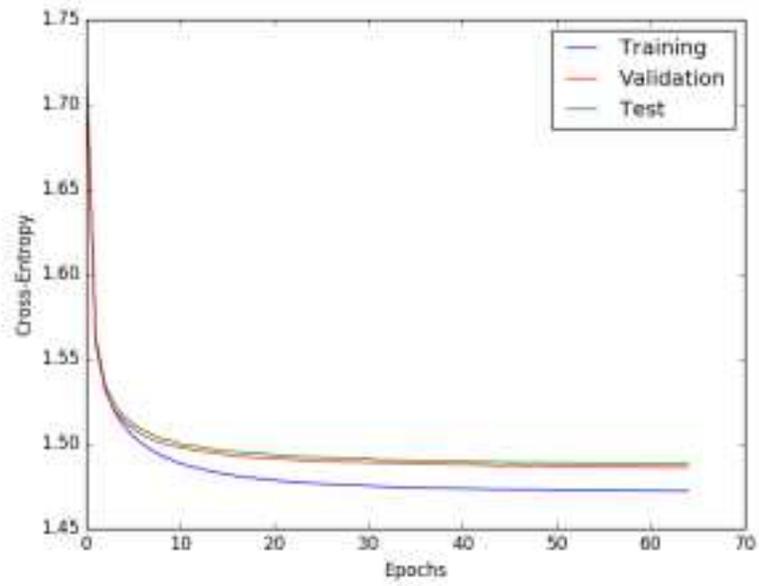}
\end{center}
\caption{Error trend on the MNIST dataset during the training of a 2-layer ANN with 300 hidden units.}\label{tfmnistloss}
 \end{figure}

\noindent The trend of the network performance showed in~\RefFig{tfmnistloss} during training can be obtained by the following code:

\begin{lstlisting}
E = range(np.shape(errors_train)[0])

line_train, = plt.plot(E, errors_train)
line_test, = plt.plot(E, errors_test)
line_val, = plt.plot(E, error_val)
plt.legend([line_train, line_val, line_test], ['Training', 'Validation', 'Test'])
plt.ylabel('Cross-Entropy')
plt.xlabel('Epochs')
plt.show()
\end{lstlisting}

\newpage

\section{Convolutional Neural Networks}

 \begin{figure}[H]
 \begin{adjustbox}{max size={\textwidth}}{}
 \includegraphics{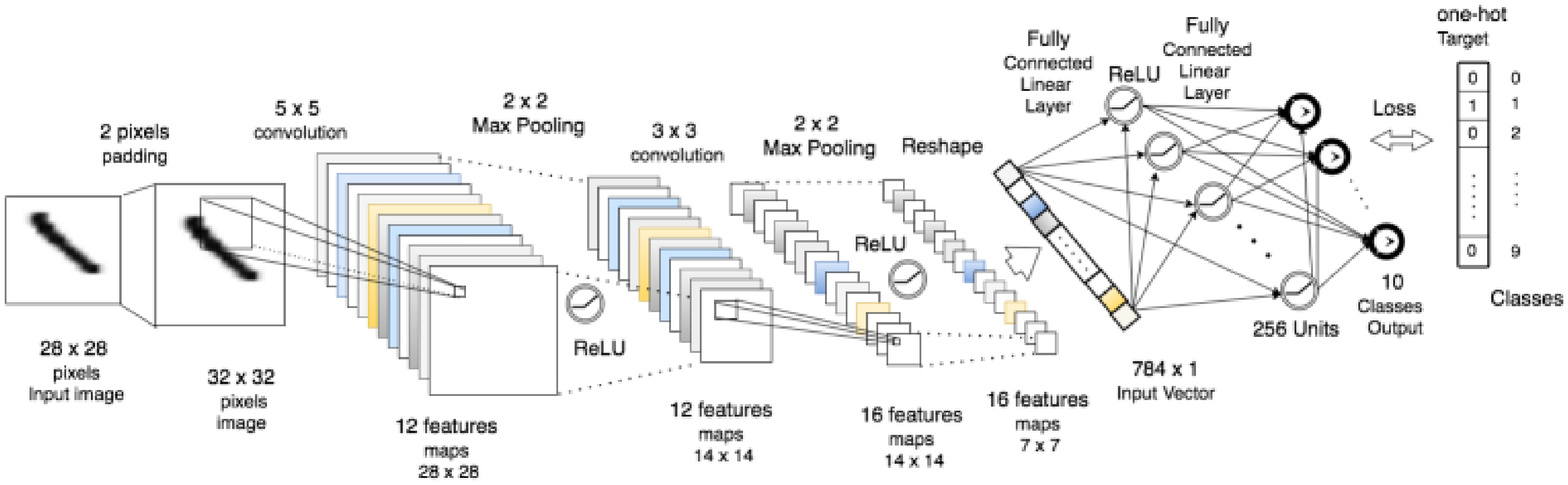}
\end{adjustbox}
\caption{General architecture of the CNN model proposed to face the MNIST data.}\label{cnnmnistmod}
 \end{figure}
 
In this section we introduce the Convolutional Neural Networks (CNNs~\cite{fukushima1988neocognitron,lecunn98,serre2007robust}), an important and powerful kind of learning architecture widely diffused especially for Computer Vision applications. They currently represent state of the art algorithm for image classification tasks and constitute the main architecture used in Deep Learning. We show how to build and train such a structure within all the proposed frameworks, exploring the most general functions and setting up few experiments on MNIST pointing out some important features.

\subsection{\MTL}\label{cnnmatsec}

\mcode
Main function and classes to build and train CNNs with \MTL are contained again in \NNtb and \SMLtb. Nevertheless, the \PCtb becomes necessary too. Moreover, to train the network a CUDA\Cpr -enabled NVIDIA\Cpr GPU is required. Again, we do not focus too much on main theoretical properties and general issues about CNNs but only on main implementation instruments. 

\noindent The network has to be stored in an \MTL object of kind \tt{Layer}, which can be sequentially composed by different sub-layers. For each one, we do not list all the available options, which as always can be explored by the interactive help options from the Command Window\footnote{Further documentation is available at the official site \url{https://it.mathworks.com/help/nnet/convolutional-neural-networks.html}}. Most common convolutional objects can be defined by the functions:

\begin{description}
\item[\tt{imageInputLayer}] creates the layer which deals with the original input image, requiring as argument a vector expressing the size of the input image given by \emph{height} by \emph{width} by  \emph{number of channels};
\item[\tt{convolution2dLayer}] defines a layer of 2-D convolutional filters whose size is specified by the first argument (a real number for a square filter, a 2-D vector to specify both height and width), whereas the second argument specifies the total number of filters; main options are \tt{'Stride'}, which indicates the sliding step (default \tt{[1, 1]} means 1 pixel in both directions), and \tt{'Padding'} (default \tt{[0, 0]}), whose have to appear in Name,Value pairs;
\item[\tt{reluLayer}] defines a layer computing the Rectifier Activation Linear Unit (ReLU) for the filter outputs;
\item[\tt{averagePooling2dLayer}] layer computing a spatial reduction of the input by averaging the values of the input on each grid of given dimension (default \tt{[2, 2]}, \tt{'Stride'} and \tt{'Padding'} are options too);
\item[\tt{maxPooling2dLayer}] layer computing a spatial reduction of the input assigning the max value to each grid of given dimensions (default \tt{[2, 2]}, \tt{'Stride'} and \tt{'Padding'} are options too);
\item[\tt{fullyConnectedLayer}]	requires the desired output dimension as argument and instantiates a classic fully connected linear layer, the number of the connections is adapted to fit the input size;
\item[\tt{dropoutLayer}] executes a dropout units selection with the probability given as argument;
\item[\tt{softmaxLayer}] computes a probability normalization based on the \emph{softmax} function;
\item[\tt{classificationLayer}] adds the final classification layer evaluating the \emph{Cross-Entropy} loss function between predictions and labels
\end{description}

\begin{lstlisting}[caption=Definition of a CNN to face the MNIST within Matlab framework.,label=asd]
CnnM = [ imageInputLayer([28 28 1]), ... % input layer
... % convolutional layer 12 filters of size 5x5 and 2 pixels 0-padding
convolution2dLayer(5,12,'Padding',2), ...
reluLayer(), ... 
... % max-pooling layer on 2x2 grid and 2 pixels step in both directions
maxPooling2dLayer(2,'Stride',2), ...
... % convolutional layer with 16 filters of size 3x3 and 1 pixels 0-padding
convolution2dLayer(3,16,'Padding',1), ...
reluLayer(), ... 
... % max-pooling layer on 2x2 grid and 2 pixels step in both directions
maxPooling2dLayer(2,'Stride',2), ...
... % classic fully connected layer with 256 output
fullyConnectedLayer(256), ...
reluLayer(), ...
... % classic fully connected layer with 10 output         
fullyConnectedLayer(10), ...
softmaxLayer(), ...
classificationLayer() ];
\end{lstlisting}
 
\noindent In \RefList{asd} we show how to set up a basic CNN to face the $28 \times 28$ pixels images from MNIST showed in~\RefFig{cnnmnistmod}. The global structure of the network is defined as a vector composed by the different modules. The initialized object \tt{Layer}, named \emph{CnnM}, can be visualized from the command window giving:
 
 \vspace{0.3cm}
\includegraphics{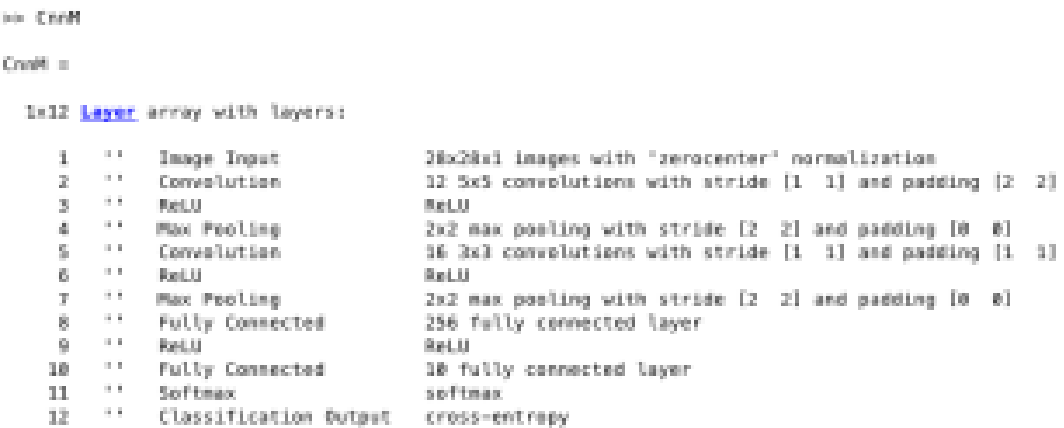}
 \vspace{0.3cm}
  
\noindent Data for the MNIST can be loaded by the Stanford routines showed in \RefSec{matmnist}. This time input images are required as a 4-D tensor of size \emph{numberOfSamples}--by--\emph{channels}--by--\emph{height}--by--\emph{width} and, hence, we have to modify the provided function \tt{loadMNISTimages} or just to reshape data, as showed in the first line of the following code:

\begin{lstlisting}
X = reshape(images_Train,28,28,1,[])
Y = nominal(labels_Train,{'0','1','2','3','4','5','6','7','8','9'});
\end{lstlisting}
 
\noindent  The second command is used to convert targets in a \tt{categorical} \MTL variable of kind \tt{nominal}, which is required to train the network exploiting the function \tt{trainNetwork}. It also require as input an object specifying the training options which can be instantiated by the function \tt{trainingOptions}. The command:
 
 \begin{lstlisting}
opts = trainingOptions('sgdm');
\end{lstlisting}

\noindent selects (by the \tt{'sgdm'} string) the \emph{Stochastic Gradient Descent} algorithm using momentum. Many optional parameters are available, which can be set by additional parameter in the \emph{Name,Value} pairs notation again. The most common are:
\small{
\begin{description}
\item[\tt{Momentum}] (default $0.9$)
\item[\tt{InitialLearnRate}] (default $0.01$)
\item[\tt{L2Regularization}] (default $0.0001$)
\item[\tt{MaxEpochs}] (default $30$)
\item[\tt{MiniBatchSize}] (default $128$)
\end{description}}

\noindent After this configuration, the training can be started by the command:
 \begin{lstlisting}
[trainedNet,trainOp] = trainNetwork(X,Y,CnnM,opts)
\end{lstlisting}

\noindent where \emph{trainedNet} will contain the trained network and \emph{trainOp} the training variables. Training starts a command line printing of some useful variables indicating the training state, which will be similar to:

 \vspace{0.3cm}
\includegraphics{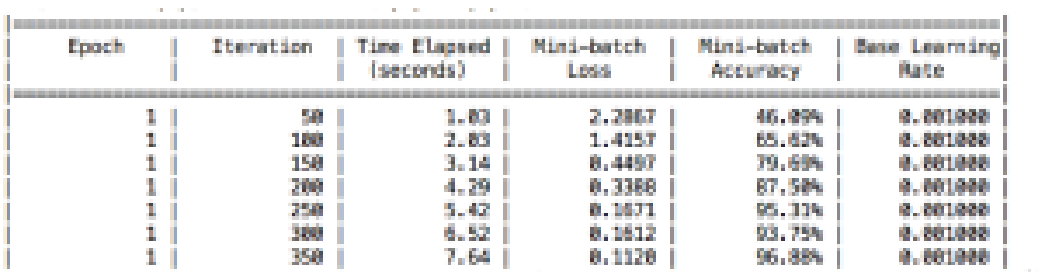}
 \vspace{0.3cm}
 
 \noindent At the end of the training, we can evaluate the performance on a suitable test set \emph{Xtest}, together with its correspondent target \emph{Ytest}, by the function \tt{classify}:
 \begin{lstlisting}
Predictions = classify(trainedNet,Xtest);
Accuracy = mean(Predictions==Ytest);
C = confusionmat(P,categorical(Y));
\end{lstlisting}

\noindent Assuming \emph{Ytest} to be a 1-Dimensional vector of class labels, classification accuracy can be calculated as before by the meaning of the boolean comparing with the computed predictions vector \emph{Predictions}. An useful built-in function to compute the confusion matrix is provided, requiring the \tt{nominal} labels to be converted into \tt{categorical} as the predictions. In this setting, the final classification accuracy on the test set should be close to the 99\%. 

\begin{figure}[H]
 	\begin{center}
 		\includegraphics[height=5 cm]{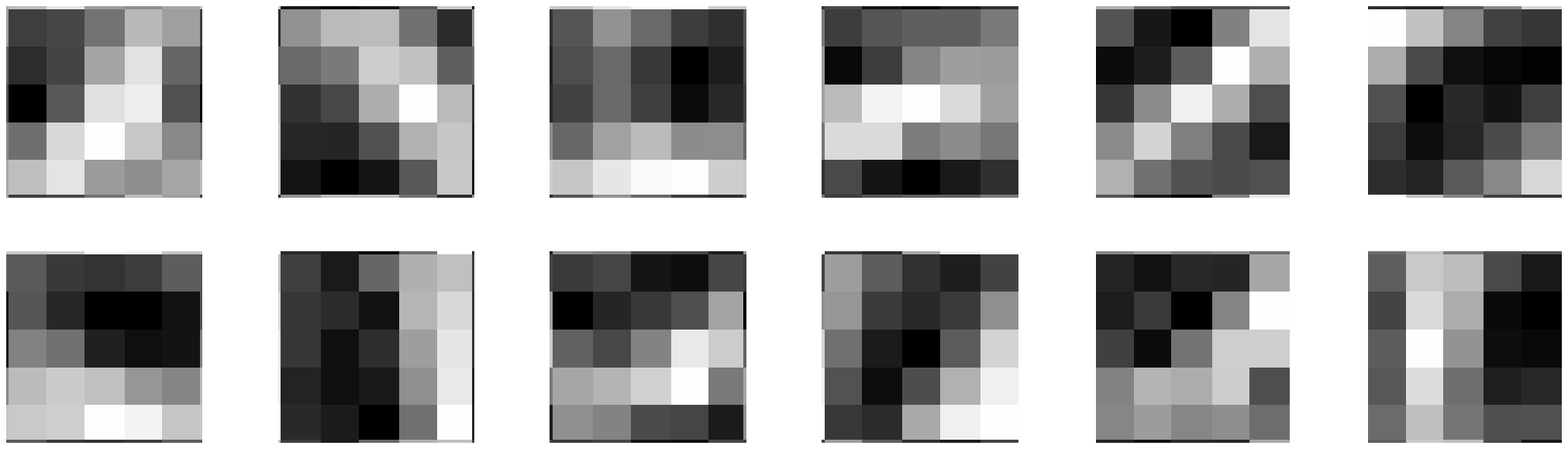}
 	\end{center}
\caption{First Convolutional Layer filters of dimension $5\times 5$ after the training on the MNIST images with \MTL.}\label{mnistmatlabfilters}
 \end{figure}

When dealing with CNNs, an important new type of object introduced are the convolutional filters. We do not want to go in deep with theoretical explanations, however sometimes it could be useful to visualize the composition of the convolutional filters in order to get an idea of which kind of features each filter detects. Filters are represented by the weights of each convolution, stored in each layers in the 4-D weights tensor of size \emph{height}--by--\emph{width}--by--\emph{numberOfChannels}--by--\emph{numberOfFilters}. In our case for example, the filters of the first convolutional  be accessed by the notation \tt{CnnM.Layer(2).Weights}. In \RefFig{mnistmatlabfilters}, we show their configuration after the training (again exploiting the function \tt{image} and the colormap \tt{gray} and a normalization in $[0,255]$ for a suitable visualization).

\subsection{Torch}\label{cnntorchsec}
\lcode
Within the Torch environment is straightforward to define a CNN from the \tt{nn} package presented in \RefSec{torchintro}. Indeed, the network can be stored in a container and composed by specific modules. Again, we give a short list description of the most common ones, which can be integrated with the standard transfer functions or with a standard linear (fully connected) layer introduced before:

\begin{description}
\item[\tt{SpatialConvolution}] defines a convolutional layers, the required arguments are the number of input channels, the number of output channels, the height and the width of the filters. The step-size and zero-padding height and width are optional parameters (with default value 1 and 0 respectively) 
\item[\tt{SpatialMaxPooling}] standard max pooling layer, requiring as inputs height and width of the pooling window, whereas the step-sizes are optional parameters (default the same as the window size)
\item[\tt{SpatialAveragePooling}] standard average pooling layer, same features of the previous one
\item[\tt{SpatialDropout}] set a dropout layer taking as optional argument the deactivating rate (default 0.5)
\item[\tt{Reshape}] is a module which is usually used to unroll the output after a convolutional/pooling process as a 1-D vector to be feed to a linear layer, takes as input the size of the desired output dimensions
\end{description}

\begin{figure}[H]
 \begin{adjustbox}{max size={\textwidth}}{}
 \includegraphics{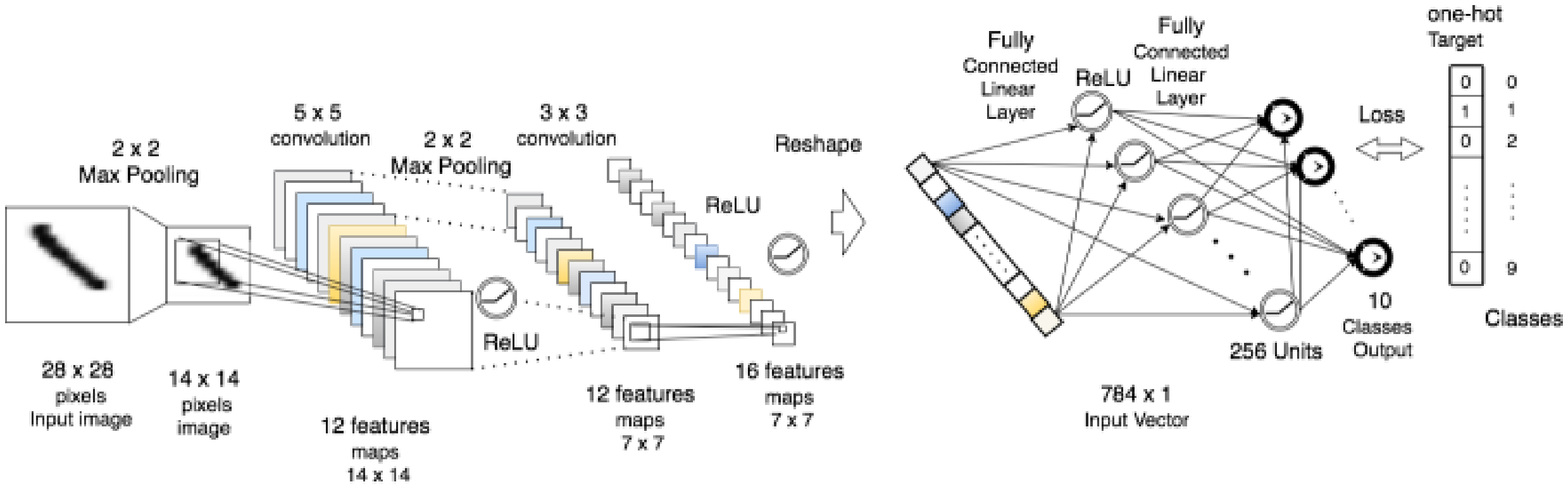}
\end{adjustbox}
\caption{General architecture of the CNN model proposed to face the MNIST data within the Torch environment (\RefSec{cnntorchsec}).}\label{cnnmnistmodtorch}
 \end{figure}

\noindent The assembly of the network follows from what seen until now. To have a different comparison with the previous experiment, we operate an initial $2\times 2$ window max-pooling on the input image, in order to provide the network by images of lower resolution. The general architecture will differ from the one defined in \RefSec{cnnmatsec} only by the first layers. The proposed network is generated by the code in \RefList{cnntorchnet}, whereas in  \RefFig{cnnmnistmodtorch} we show the global architecture.

\begin{lstlisting}[caption=Network definition for MNIST data reduced to $14 \times 14$ pixels images by $2\times 2$ max-pooling, label=cnntorchnet]
require 'nn'  
-- container definition
cnnet = nn.Sequential()
-- network assembly
cnnet:add(nn.SpatialConvolution(1,12,5,5,1,1,2,2))
cnnet:add(nn.ReLU())
cnnet:add(nn.SpatialMaxPooling(2,2,2,2)) 
cnnet:add(nn.SpatialConvolution(12,16,3,3,1,1,1,1))
cnnet:add(nn.ReLU())
cnnet:add(nn.Reshape(7*7*16))
cnnet:add(nn.Linear(7*7*16,256))
cnnet:add(nn.ReLU()) 
cnnet:add(nn.Linear(256,10))
cnnet:add(nn.SoftMax()) 
\end{lstlisting}

\begin{figure}[H]
 	\begin{center}
 		\includegraphics[height=5 cm]{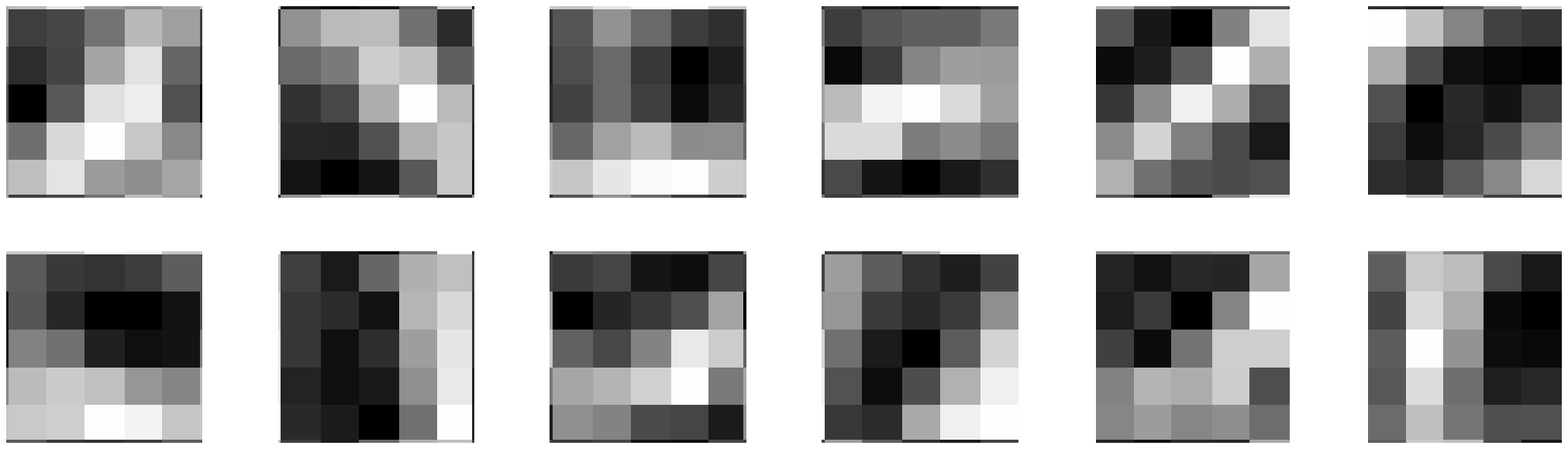}
 	\end{center}
\caption{First Convolutional Layer filters of dimension $5\times 5$ after the training with Torch on the MNIST images halved by $2\times 2$ max pooling.}\label{mnisttorchfilters}
 \end{figure}

\noindent If we use the training function defined in \RefList{mnisttorchtrain}, the optimization (starting with the same options) stops for validation check after 123 epochs, producing a Classification Accuracy of about 90\%. This just to give an idea of the difference in the obtained performances when there is a reduction in the information expressed by input images. In \RefFig{mnisttorchfilters} we show the 12 filters of size $5\times 5$ extracted by the first convolutional layer. The weights can be obtained by the function \tt{parameters} from the package \tt{nn}:

 \begin{lstlisting}
myParam = cnnet:parameters()
\end{lstlisting}

\noindent which return an indexed table storing the weights of each layer. In this case, the first element of the table contains a tensor of dimension 12 by 25 representing the weights of the filters. The visualization can be generated exploiting the function \tt{imagesc} from the package \tt{gnuplot}, after reshaping each line in the $5\times 5$ format. 

\subsection{Tensor Flow}
In this section we will show how to build a CNN to face the MNIST data using TensorFlow. The first step is to import libraries, the Mnist Dataset and to define main variables. Two additional package are required: \tt{numpy} for matrices computations and \tt{matplotlib.pyplot} for visualization issues.

\pcode
\begin{lstlisting}
import tensorflow as tf	
import numpy as np
import matplotlib.pyplot as plt
	
from tensorflow.examples.tutorials.mnist import input_data
mnist = input_data.read_data_sets('MNIST_data', one_hot=True)
	
x = tf.placeholder(tf.float32, shape=[None, 784])
y_ = tf.placeholder(tf.float32, shape=[None, 10])
\end{lstlisting}

\noindent We define now some tool functions to specify variable initialization.

\pcode
\begin{lstlisting}
def weight_variable(shape):
	initial = tf.truncated_normal(shape, stddev=0.1)
	return tf.Variable(initial)
	
def bias_variable(shape):
	initial = tf.constant(0.1, shape=shape)
	return tf.Variable(initial)
\end{lstlisting}

\noindent The following two function define convolution and (3-by-3) max pooling. The vector \emph{strides} specifies how the filter or the sliding window move along each dimension. The vector \emph{ksize} specifies the dimension of the sliding window. The padding option \emph{'SAME'} automatically adds empty (zero valued) pixels to allow the convolution to be centered even in the boundary pixels.

\pcode
\begin{lstlisting}
def conv2d(x, W):
	return tf.nn.conv2d(x, W, strides=[1, 1, 1, 1], padding='SAME')

# max pooling over 3x3 blocks
def max_pool_3x3(x):
	return tf.nn.max_pool(x, ksize=[1, 3, 3, 1],
			strides=[1, 3, 3, 1], padding='SAME')
\end{lstlisting}

\noindent In order to define the model, we start by reshaping the input (where each sample is provided as 1-D vector) to its original size, i.e. each sample is represented by a matrix of 28x28 pixels. Then we define the first convolution layer which computes 12 features by using 5x5 filters. Finally we perform the \tt{ReLU} activation and the first max pooling step.

\begin{lstlisting}
# Input resize
x_image = tf.reshape(x, [-1,28,28,1]) 

# First convolution layer - 5x5 filters
INPUT_C1 = 1 # input channel
OUTPUT_C1 = 12 # output channel (features)
W_conv1 = weight_variable([5, 5, INPUT_C1, OUTPUT_C1])
b_conv1 = bias_variable([OUTPUT_C1])

#convolution step
h_conv1 = tf.nn.relu( conv2d(x_image, W_conv1) + b_conv1 )

#max pooling step
h_pool1 = max_pool_3x3(h_conv1)
\end{lstlisting}

\noindent The second convolution layer can be built up in an analogous way.

\begin{lstlisting}
# Second convolution layer
INPUT_C2 = OUTPUT_C1
OUTPUT_C2 = 16 # output channel (features)
W_conv2 = weight_variable([5, 5, INPUT_C2, OUTPUT_C2])
b_conv2 = bias_variable([OUTPUT_C2])
	
#convolution step
h_conv2 = tf.nn.relu(conv2d(h_pool1, W_conv2) + b_conv2)
	
#max pooling step
h_pool2 = max_pool_3x3(h_conv2)
\end{lstlisting}

\noindent At this point the network returns 16 feature maps 4x4. These will be reshaped to 1--D vectors and given as input to the last fully connected linear layer. The linear layer is equipped with 1024 hidden units with \emph{ReLU} activation functions.\\
\\
\begin{lstlisting}
# Definition of the fully connected linear layer
FS = 4 # final size 
W_fc1 = weight_variable([ FS * FS * OUTPUT_C2, 1024])
b_fc1 = bias_variable([1024])
	
# Reshape images
h_pool2_flat = tf.reshape(h_pool2, [-1, FS * FS * OUTPUT_C2])
	
# hidden layer
h_fc1 = tf.nn.relu(tf.matmul(h_pool2_flat, W_fc1) + b_fc1)
	
# Output layer
W_fc2 = weight_variable([1024, 10])
b_fc2 = bias_variable([10])
	
# Prediction
y_conv = tf.matmul(h_fc1, W_fc2) + b_fc2
\end{lstlisting}

\noindent In the following piece of code we report the optimization process of the defined CNN. As in the standard ANN case, it is organized in a \tt{for} loop. This time, we chose the \emph{Adam} gradient-based optimization by the function \tt{AdamOptimizer}. Each epoch performs a training step over mini-batches extracted again by the dedicated function \tt{next\_batch()}, introduced in~\RefSec{tfmnist}. This time the computations are run within \emph{InteractiveSession}. The difference with the regular \emph{Session} is that an \emph{InteractiveSession} sets itself as the default session during building, allowing to run variables without needing to constantly refer to the session object. As a for instance, the method \tt{eval()} will implicitly use that session to run operations.

\begin{lstlisting}
cross_entropy = tf.reduce_mean(tf.nn.softmax_cross_entropy_with_logits(y_conv, y_))
train_step = tf.train.AdamOptimizer(1e-4).minimize(cross_entropy)
correct_prediction = tf.equal(tf.argmax(y_conv,1), tf.argmax(y_,1))
accuracy = tf.reduce_mean(tf.cast(correct_prediction, tf.float32))
	
sess = tf.InteractiveSession()
sess.run(tf.global_variables_initializer())

# Early stopping setup, to check on validation set
prec_err = 10**6 # just a very big value
val_count = 0
val_max_steps = 6

epochs = 100
batch_size = 1000
num_of_batches = 60000/batch_size

i=1
while i <= epochs and val_count < val_max_steps:

	print 'Epoch:',i,'(Early stopping criterion:',val_count,'/',val_max_steps,')'

	# training step
	for j in range(num_of_batches):
		batch = mnist.train.next_batch(batch_size)
		sess.run(train_step, feed_dict={x: batch[0], y_: batch[1]})
	
	# visualize accuracy each 10 epochs
	if i == 1 or i%10 == 0:
		train_accuracy = accuracy.eval(
					feed_dict={x: mnist.train.images, y_: mnist.train.labels})    
		test_accuracy = accuracy.eval(
					feed_dict={x: mnist.test.images, y_: mnist.test.labels})
		print "\nAccuracy at epoch ", i, ":"
		print("train accuracy %g, test accuracy %g\n"%(train_accuracy, test_accuracy))
	
	# validation check
	curr_err = sess.run(cross_entropy, 
				feed_dict={x: mnist.validation.images, y_: mnist.validation.labels})
	if curr_err >= prec_err*0.9999:
		val_count = val_count + 1
	else:
		val_count = 0
	prec_err = curr_err
	
	i+=1


print("\n\nResult:\nTest accuracy %g" % accuracy.eval(
			feed_dict={x: mnist.test.images, y_: mnist.test.labels}))	
	
	
\end{lstlisting}

\begin{figure}[H]
 \begin{center}
\includegraphics[height=7 cm]{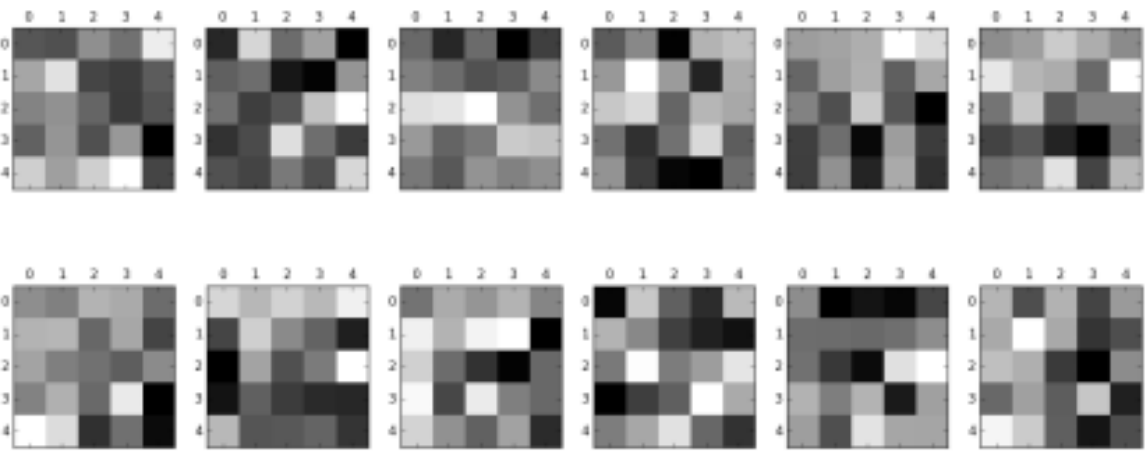}
\end{center}
\caption{First Convolutional Layer filters of dimension $5\times 5$ after the training with TensorFlow on the MNIST images with the described arcitecture.}\label{tfcnnfilters}
 \end{figure}

\noindent In this setting, the final classification accuracy on the test set should be close to the 99\%. As in the previous Sections, in~\RefFig{tfcnnfilters} we show the learned filters of the first convolutional layer obtained by:

\begin{lstlisting}
# Getting filters as an array
FILTERS = W_conv1.eval()
	
fig = plt.figure()
for i in range(np.shape(FILTERS)[3]):
	ax = fig.add_subplot(2, 6, i+1)
	ax.matshow(FILTERS[:,:,0,i], cmap='gray')
plt.show()
\end{lstlisting}

In the first line \tt{eval()} computes the current value of \tt{W\_conv1}, saving it in the 4--D \tt{numpy} array \emph{FILTERS}, where the fourth dimension (acceded by the index 3 in \tt{np.shape(FILTERS)[3]}) corresponds to the number of filters.

\newpage

\section{A critical comparison}  

In this Section we would like to outline an overall picture across the presented environments. Even if in \RefTab{comptab} we provide a scoring based on some features we thought mainly relevant for Machine Learning software development, this work would not like to bound this analysis to a poor evaluation. Instead, we hope to propose an useful guideline to help people trying to approach ANNs and Machine Learning in general, in order to orientate within the environments depending on personal background and requirements. More complete and statistically relevant comparisons can be found on the web\footnote{Look for example at the webpage \url{http://hammerprinciple.com/therighttool}}, but we try to summarize so as to help and speed up single and global task developing.

We first give general description of each environment, then we try to compare pros and cons on specific requirements. At the end, we carry out an indicative numerical analysis on the computational performances on different tasks, which could be also a topic for comparison and discussion.

\subsection{\MTL}

The programming language is intuitive and the software provides a complete package, allowing the user to define and train almost all kind of ANNs architecture without writing a single line of specific code. The code parallelization is automatic and the integration with CUDA\Cpr is straightforward too.  The available built-in functions are very customizable and optimized, providing fast and extended setting up of experiments and an easy access to the variable of the network for in-depth analysis. However, enlarging and integrating \MTL tools require an advanced knowledge of the environment. This could drive the user to start rewriting its own code from the beginning, leading to a general decay of computational performances. These features make it perfect as a statistical and analysis toolbox, but maybe a bit slow as developmental environment. The GUI results sometimes heavy to be handled by the calculator, but, on the other hand, it is very user-friendly and provides the best graphical data visualization. The documentation is complete and well organized within the official site.

\subsection{Torch}

The programming language (Lua) can sometimes results a little bit tricky, but it supposed to be the faster among these languages. It provides all the needed CUDA\Cpr integrations and the CPU parallelization automatic. The module-based structure allows flexibility in the ANNs architecture and it is relatively easy to extend the provided packages. There are also other powerful packages\footnote{We skip the treatment of \tt{optim}, which provides various Gradient Descent and Back-Propagation procedure},  but in general they require to acquire some expertise to achieve a conscious handling.
Torch could be easily used as a prototyping environment for specific and generic algorithms testing. The documentation is spread all over the \tt{torch} GitHub repository and sometimes solve specific issues could not be immediate.

\subsection{Tensor Flow}

The employment of a programming language as dynamic as Python makes the code scripting light for the user. The CPU parallelization is automatic, and, exploiting the graph-structure of the computation is easy to take advantage of GPU computing. It provides a good data visualization and the possibility for beginners to access to ready to go packages, even if not treated in this document. The power of symbolic computation involves the user only in the forward step, whereas the backward step is entirely derived by the TensorFlow environment. This flexibility allows a very fast development for users from any level of expertise.

\newpage
\subsection{An overall picture on the comparison}

As already said, in \RefTab{comptab} we try to sum up a global comparison trying assigning a score from 1 to 5 on different perspectives. Here below, we explain the main motivation when necessary:

\begin{description}
\item[Programming Language] All the basic language are quite intuitive.
 \item[GPU Integration] \MTL is penalized since an extra toolbox is required.
 \item[CPU Parallelization] All the environments exploit as more core as possible
 \item[Function Customizability] \MTL score is lower since integrate well-optimized functions with the provided ones is difficult
\item[Symbolic Calculus] Not expected in Lua
\item[Network Structure Customizability] Every kind of network is possible
\item[Data Visualization] The interactive \MTL mode outperforms the others
\item[Installation] Quite simple for all of them, but the \MTL interactive GUI is an extra point
\item[OS Compatibility] Torch installation is not easy on Windows
\item[Built-In Function Availability] \MTL provided simple-tools with an easy access
\item[Language Performance ] \MTL interface can sometimes appear heavy
\item[Development Flexibility] Again, \MTL is penalized because it forces medium users to become very specialized with the language to integrate the provided tools or to write proper code, which in general can slow down the software development
\end{description}

\vfill
\begin{table}[H]
\begin{center}
	\begin{tabular}{ | l | l | l | p{5cm} |}
		\hline
		 & \MTL & Torch & TensorFlow \\ \hline
		Programming Language & 4 & 3 & 4 \\ \hline
		GPU Integration & 3 & 5 & 5 \\ \hline
		CPU Parallelization & 5 & 5 & 5 \\ \hline
		Function Customizability & 2 & 4 & 5 \\ \hline
		Symbolic Calculus & 3 & 1 & 5 \\ \hline
		Network Structure Customizability  & 5 & 5 & 5 \\ \hline
		Data Visualization & 5 & 2 & 3 \\ \hline
		Installation & 5 & 4 & 4 \\ \hline
		OS Compatibility & 5 & 4 & 5 \\ \hline
		Built-In Function Availability & 5 & 4 & 4 \\ \hline
		Language Performance & 3 & 5 & 4 \\ \hline
Development Flexibility & 2 & 4 & 5 \\ \hline
		License & EULA & BSD (Open source) & Apache 2.0 (Open source) \\ \hline
	\end{tabular}
\end{center}
\caption{Environments individual scoring}\label{comptab}
\end{table}

\newpage

\subsection{Computational issues} 

In \RefTab{perftab} we compare running times for different tasks, analyzing the advantages and differences of CPU/GPU computing. Results are averaged on 5 trials, carried out in the same machine with an Intel(R) Xeon(R) CPU E5-2650 v2 @ 2.60GHz with 32 cores, 66 GB of RAM, and a Geforce GTX 960 with 4GB of memory. The OS is Debian GNU/Linux 8 (jessie). 
We test a standard Gradient Descent procedure varying the network architecture, the batches size (among \emph{Stochastic Gradient Descent} (SGD), 1000 samples batch and Full Batch) and the hardware (indicated in HW column). The CNN architecture is the same of the one proposed in~\RefFig{cnnmnistmod}. Performances are obtained trying to use optimization procedures as similar as possible. In practice, it is very difficult to reply the specific optimization techniques exploited in \MTL built-in toolboxes. We skip the SGD case for the second architecture (eighth row) in Torch because of the huge computational time obtained for the first architecture. We miss the SGD case for the ANNs architecture in the \MTL case since the training function \emph{'trains'} it is not supported for GPU computing (rows fourth and tenth). As a matter of fact, this could be an uncommon case of study, but we report the results for best completeness. We skip the CNN Full Batch trials on GPU because of the too large memory requirement\footnote{For an exhaustive comparison on computational performance on several tasks (including the comparison with other existent packages) the user can refer to \url{https://en.wikipedia.org/wiki/Comparison_of_deep_learning_software}} .

\vfill
\begin{table}[H]
\begin{center}
{\tiny
\resizebox{\textwidth}{!}{%
\begin{tabu}{|c|c|c|c|c|c|}
\hline
  & \multicolumn{3}{c|}{\rule{0pt}{3ex} Batches size } & & \\
  \cline{2-4}
  \multirow{-2}{*}{Architecture} & \rule{0pt}{3ex}  SGD   & 1000 & Full Batch&\multirow{-2}{*}{Env.} & \multirow{-2}{*}{HW}\\
 \tabucline[2pt]{-}
   \rowcolor{gray!5}  \cellcolor{white!50} \rule{0pt}{3ex}  &  52.46  &   30.40 &  28.38   & Matlab &  \cellcolor{white!50} \\ 
   \rowcolor{gray!10}  \cellcolor{white!50} \rule{0pt}{3ex}  &    3481.00    & 46.55 & 24.43 & Torch &  \cellcolor{white!50} \\ 
    \rowcolor{gray!20}  \cellcolor{white!50} \rule{0pt}{3ex}  2-Layers ANN& 914.56 & 13.82 & 12.00 &  TF &  \cellcolor{white!50}  \multirow{-3}{*}{CPU} \\ 
\arrayrulecolor{black}\cline{2-6}
      \rowcolor{gray!5}  \cellcolor{white!50} \rule{0pt}{3ex} 1000 HUs & --  &3.75   &3.73  & &  \cellcolor{white!50} \\ 
   \rowcolor{gray!15}  \cellcolor{white!50} \rule{0pt}{3ex}  & 378.19 &  1.93 & 1.46  &  &  \cellcolor{white!50} \\ 
    \rowcolor{gray!25}  \cellcolor{white!50} \rule{0pt}{3ex} &  911.32 & 5.40 & 4.92 &  & \cellcolor{white!50}  \multirow{-3}{*}{GPU}\\ 
 \tabucline[1pt]{-}    
      \rowcolor{gray!5}  \cellcolor{white!50} \rule{0pt}{3ex}   &  44.66  & 28.33 &  27.07 & &  \cellcolor{white!50} \\ 
   \rowcolor{gray!15}  \cellcolor{white!50} \rule{0pt}{3ex}  & --  & 52.99 & 20.00 &  &  \cellcolor{white!50} \\ 
    \rowcolor{gray!25}  \cellcolor{white!50} \rule{0pt}{3ex}  4-Layers ANN& 893.27 & 10.95 & 8.83 &  &  \cellcolor{white!50} \multirow{-3}{*}{CPU}  \\ 
\arrayrulecolor{black}\cline{2-6}
      \rowcolor{gray!5}  \cellcolor{white!50} \rule{0pt}{3ex}  300-300-300 HUs & --   & 2.74  & 3.44 & &  \cellcolor{white!50}  \\ 
   \rowcolor{gray!15}  \cellcolor{white!50} \rule{0pt}{3ex}  & 517.87  &  1.51  &  1.08  &  &  \cellcolor{white!50} \\ 
    \rowcolor{gray!25}  \cellcolor{white!50} \rule{0pt}{3ex} & 1024.29 &   4.73 & 4.27 &   &  \cellcolor{white!50} \multirow{-3}{*}{GPU}\\ 
 \tabucline[1pt]{-}   
      \rowcolor{gray!5}  \cellcolor{white!50} \rule{0pt}{3ex}  &  7794.33  & 54.21  & -- & &  \cellcolor{white!50} \\ 
   \rowcolor{gray!15}  \cellcolor{white!50} \rule{0pt}{3ex}  & 647.75 &  100.06  & -- &  &  \cellcolor{white!50} \\ 
    \rowcolor{gray!25}  \cellcolor{white!50} \rule{0pt}{3ex} \multirow{-3}{*}{CNN} & 1850.30 & 20.22 & -- &   &  \cellcolor{white!50}  \multirow{-3}{*}{GPU} \\
    \hline
\end{tabu}}}
\end{center}
\caption{{\normalsize Averaged time (in seconds) on 5 running of 10 epochs training with different architectures on the MNIST data within the presented environments. All the architectures are built up using the \emph{ReLU} as activation, the \emph{softmax} as output function and the \emph{Cross-Entropy} penalty.}}
\label{perftab}
\end{table}

\newpage


\newpage

\bibliographystyle{unsrt}
\bibliography{references}

\end{document}